\documentclass{article}
\usepackage{iclr2026_conference,times}

\usepackage[utf8]{inputenc}
\usepackage{hyperref}
\usepackage{url}
\usepackage{booktabs}
\usepackage{amsfonts}
\usepackage{nicefrac}
\usepackage{microtype}
\usepackage{xcolor}
\usepackage{amsthm}
\usepackage{amsmath}
\usepackage{enumitem}
\usepackage[table,xcdraw]{xcolor}
\usepackage{tabularx}
\usepackage{tabularx}
\usepackage{multirow}
\usepackage{amsmath}
\usepackage{graphicx}
\usepackage{tikz}
\usetikzlibrary{calc, shadows}
\usepackage{tcolorbox}
\tcbuselibrary{skins, breakable, raster, listingsutf8}
\usepackage{amssymb} 
\usetikzlibrary{positioning, arrows.meta}

\title{Ice Cream Doesn't Cause Drowning: Benchmarking LLMs Against Statistical Pitfalls in Causal Inference}

\author{
\textbf{Jin Du$^{1}$, Li Chen$^{1}$, Xun Xian$^{1}$, An Luo$^{1}$, Fangqiao Tian$^{1}$, Ganghua Wang$^{2}$,}\\
\textbf{Charles Doss$^{1}$, Xiaotong Shen$^{1}$, Jie Ding$^{1}$}\\
$^{1}$School of Statistics, University of Minnesota, Minneapolis, MN 55455\\
$^{2}$Data Science Institute, University of Chicago, Chicago, IL 60637\\
{\footnotesize\texttt{\{du000142,chen7019,xian0044,luo00318,tian0257\}@umn.edu}}\\
{\footnotesize\texttt{\{cdoss,xshen,dingj\}@umn.edu}},
\texttt{ganghua@uchicago.edu}
}

\iclrfinalcopy
\begin{document}

\maketitle

\begin{abstract}
Reliable causal inference is essential for making decisions in high-stakes areas like medicine, economics, and public policy. However, it remains unclear whether large language models (LLMs) can handle rigorous and trustworthy \textit{statistical causal inference}. Current benchmarks usually involve simplified tasks. For example, these tasks might only ask LLMs to identify semantic causal relationships or draw conclusions directly from raw data. As a result, models may overlook important statistical pitfalls, such as Simpson’s paradox or selection bias. This oversight limits the applicability of LLMs in the real world.
To address these limitations, we propose \textbf{CausalPitfalls}, a comprehensive benchmark designed to rigorously evaluate the capability of LLMs in overcoming common causal inference pitfalls. Our benchmark features structured challenges across multiple difficulty levels, each paired with grading rubrics. This approach allows us to quantitatively measure both causal reasoning capabilities and the reliability of LLMs' responses. We evaluate models using two protocols: (1) direct prompting, which assesses intrinsic causal reasoning, and (2) code-assisted prompting, where models generate executable code for statistical analysis. Additionally, we validate the effectiveness of this judge by comparing its scoring with assessments from human experts.
Our results reveal significant limitations in current LLMs when performing statistical causal inference. The {CausalPitfalls} benchmark provides essential guidance and quantitative metrics to advance the development of trustworthy causal reasoning systems. Our code is publicly available at   \href{https://github.com/dudududuu/CausalPitfalls}{\textcolor{blue}{CausalPitfalls}}.
\end{abstract}

\section{Introduction}

Causal inference~\citep{pearl2009causality,imbens2015causal} is fundamental to decision-making across diverse fields. For instance, accurately determining the effectiveness and safety of a vaccine is pivotal in public health decisions~\citep{voysey2021safety}. However, identifying causal relationships with both reliability and interpretability remains challenging. In practice, individuals without formal statistical training frequently fall into subtle pitfalls, leading to plausible yet incorrect conclusions. A classic illustration is the erroneous conclusion that ice cream sales cause drowning incidents --- overlooking the hidden confounder of hot weather causing both events~\citep{pearl2009causality,greenland1986identifiability,rosenbaum1987sensitivity}.

Given these complexities, automated tools like large language models (LLMs) present promising avenues, demonstrated by their effectiveness in scientific problem-solving~\citep{lewkowycz2022solving,achiam2023gpt,lin2025spike,lin2025spatial} and clinical reasoning~\citep{singhal2023large,yu2025multimodal}. Recent studies~\citep{wang2024causalbench, dhawan2024end,liu2024llms} have evaluated LLMs' abilities to evaluate accuracy in causal-effect estimation, but these benchmarks often neglect crucial aspects like robustness, interpretability, and susceptibility to common causal pitfalls. As a result, LLMs can produce seemingly convincing yet misleading causal conclusions.  

To illustrate why reliability assessment is crucial, we highlight two representative failure modes (detailed in Section~\ref{sec:two_pitfalls}). First, LLMs can ignore strong data evidence, but in favor of superficial semantic cues: in a synthetic health scenario with identical datasets, LLMs concluded the drink was beneficial when labeled ``HealthPlus'' and harmful when labeled ``UltraSugar,'' even when the data indicated the \textbf{opposite}. Second, LLMs can mistake random variation for genuine causal structure: when tested on research funding data from the Netherlands, LLMs attributed differences in success rates to gender bias or Simpson’s paradox, despite statistical analyses showing that neither claim is supported. These cases demonstrate that LLMs may produce confident causal claims directly contradicted by the data, showing the need for benchmarks that assess causal inference \textit{reliability}.

\begin{figure}[!htb]
  \centering
  \includegraphics[width=\linewidth]{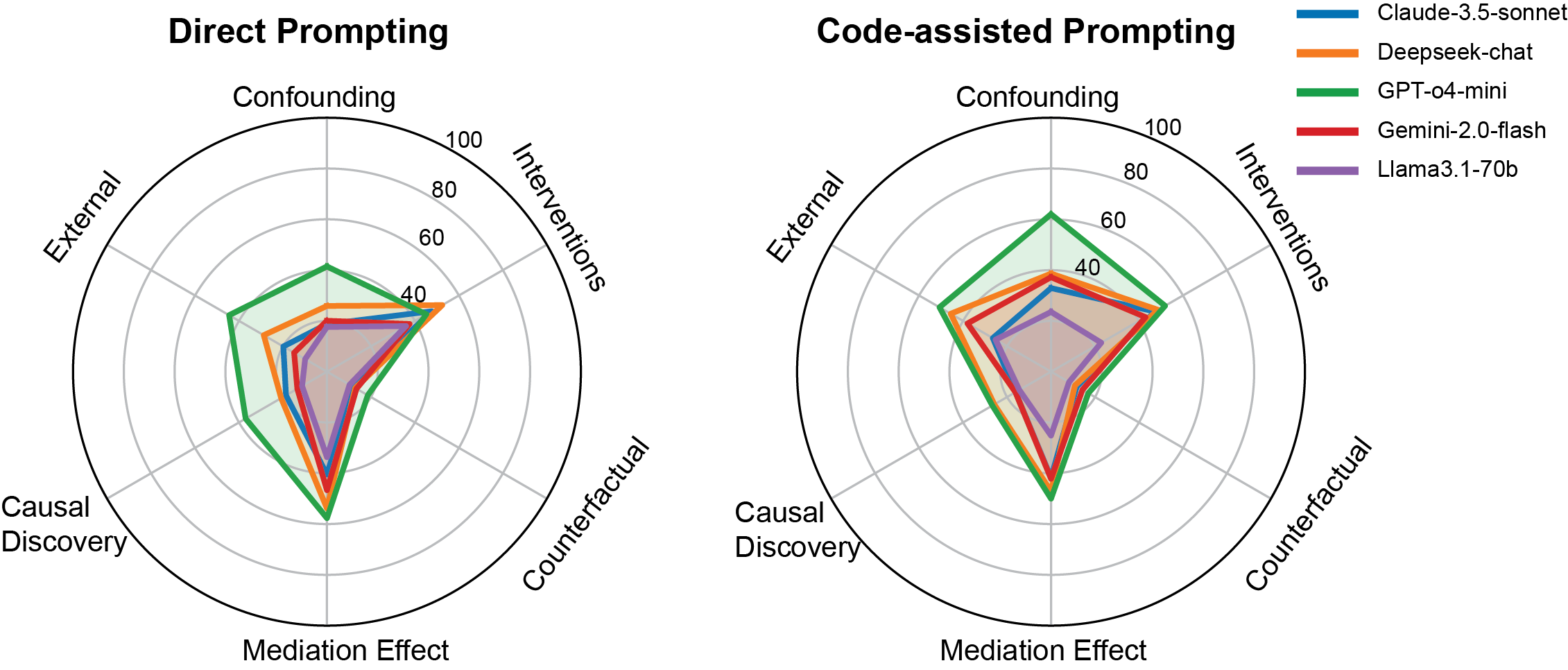}
  \caption{\textbf{Overall Message:} Our results reveal a clear \textbf{reliability gap} in causal inference when LLMs rely only on direct prompting, with all models struggling most on mediation and external validity questions. Introducing code-assisted prompting leads to substantial gains across every task and brings all models closer together in performance. This shows that executable analysis is essential for large language models to handle complex statistical challenges and deliver trustworthy causal conclusions.    
  Full results for all evaluated LLMs are provided in Table~\ref{tab:overall_performance}.
}
  \label{fig:radar_plots_intro_section}
\end{figure}

\vspace{0.2cm}
\subsection{Main contributions}
First, we introduce \textbf{CausalPitfalls}, a novel comprehensive benchmark specifically designed to evaluate the reliability of large language models (LLMs) in statistical causal inference. Unlike existing benchmarks primarily focused on accuracy, our benchmark  targets model susceptibility to common causal pitfalls as shown in Figure~\ref{fig:radar_plots_intro_section}, including (1) Confounding Biases and Spurious Associations, (2) Interventions and Experimental Reasoning, (3) Counterfactual Reasoning and Hypotheticals, (4) Mediation and Indirect Causal Effects, (5) Causal Discovery and Structure Learning, and (6) Causal Generalization and External Validity~\citep{pearl2000models,peters2017elements}. These categories are structured into 15 distinct challenges, encompassing a total of 75 evaluation questions and 75 carefully constructed datasets that systematically test the robustness of LLM causal reasoning capabilities.

Second, we comprehensively evaluate the reliability of ten LLMs under two distinct evaluation protocols: (1) \textit{direct prompting}, assessing intrinsic causal reasoning from raw data, and (2) \textit{code-assisted prompting}, where models generate executable code to perform statistical analyses before responding. This dual-protocol approach provides a detailed quantitative assessment, highlighting areas where computational assistance significantly improves causal reasoning and where intuitive reasoning could suffice.

Third, we introduce a quantitative metric termed \textit{causal reliability}, calculated as the average normalized score across all benchmark challenges, enabling standardized comparisons of LLM reliability in causal reasoning tasks. By systematically quantifying reliability, this metric provides a crucial framework for future research aimed at developing more robust and trustworthy causal inference capabilities in AI systems.

\begin{figure}[tb]
    \centering
    \includegraphics[width=0.9\linewidth]{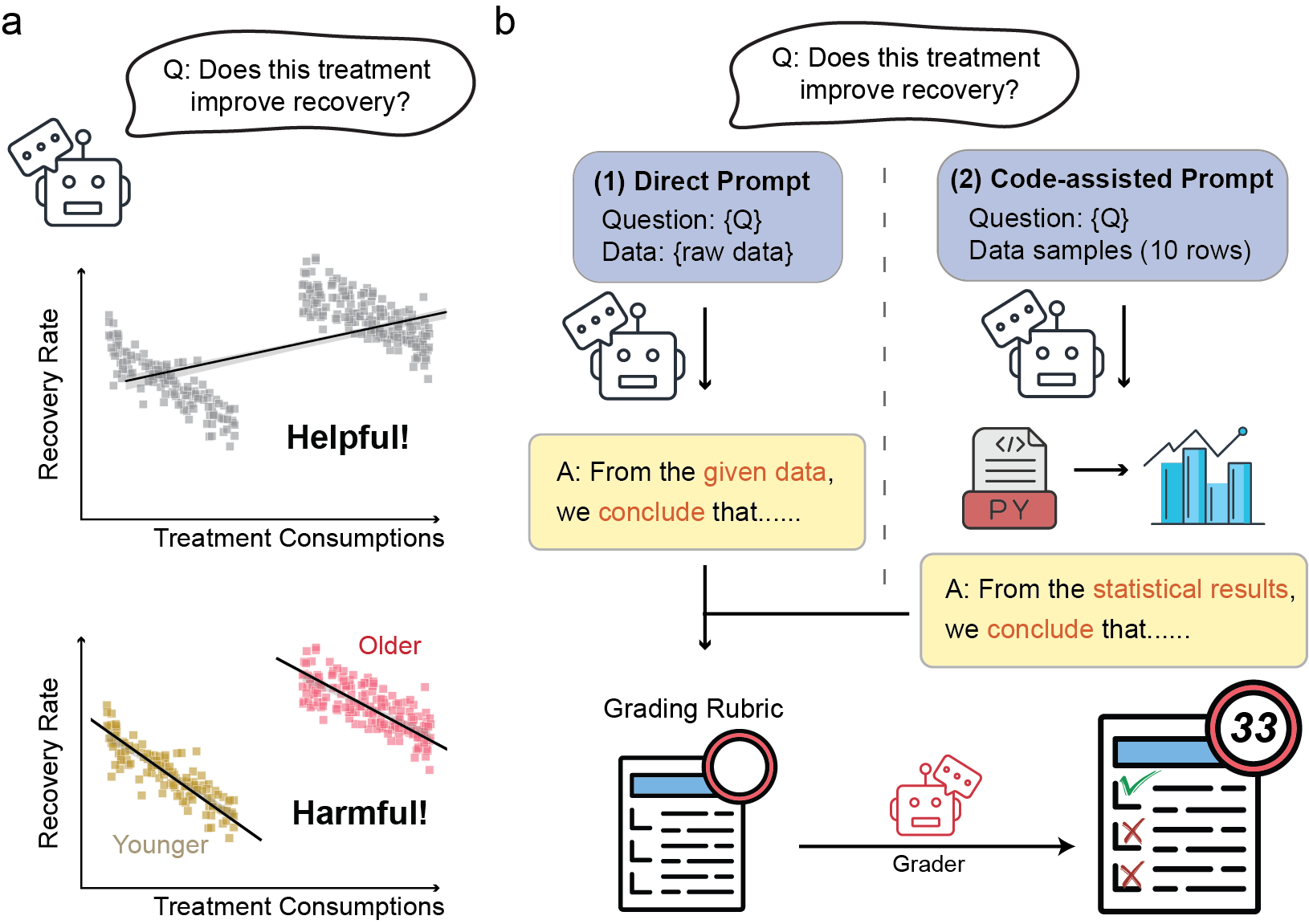}
    \caption{High-level overview of the CausalPitfalls benchmark.
(a) An illustrative real-world pitfall (Simpson’s paradox): when data on treatment consumption and recovery are pooled (top), a na{\"\i}ve analysis finds a positive effect (``Helpful!''), but stratifying by age reveals a negative effect within both younger and older subgroups (``Harmful!'').
(b) Benchmark workflow: LLMs are evaluated under two protocols: (1) Direct Prompting on raw data, assessing intrinsic causal reasoning, and (2) Code-Assisted Prompting on sampled data, assessing computationally grounded inference. In both cases, model answers are automatically scored against a hidden grading rubric by an independent grader to quantify each model’s causal reliability. }
    \label{fig:main_fig}
\end{figure}

\newpage
\subsection{Related Work}

\textbf{1. Causal Inference and Statistical Pitfalls.} 
Causal inference from observational data is inherently challenging because counterfactuals are unobservable and confounding is ubiquitous~\citep{pearl2009causality,imbens2015causal}. Causal inference methods for addressing confounders, whether the confounders are measured~\citep{chan2016globally,lin2023estimation,doss2024nonparametric} or latent~\citep{kang2016instrumental,guo2022doubly}, depend on restrictive model assumptions that are often difficult to verify empirically. Inferring causal direction similarly hinges on stringent structural equation assumptions~\citep{peters2014identifiability,li2024nonlinear} or auxiliary information such as valid instruments~\citep{chen2024discovery}. Mediation analysis~\citep{mackinnon2012introduction,yang2021estimation}, which targets specific causal pathways, demands careful adjustment for intermediate variables to avoid post‐treatment bias. Finally, transporting causal conclusions across different domains requires justification of source-target invariances and methodologies for causal knowledge transfer~\citep{wei2023transfer,chen2025enhancing}. Rigorously confronting each of these challenges is essential to conduct a reliable causal analysis.

\textbf{2. LLMs for Causal Reasoning.} 
Recent studies have extensively investigated the causal reasoning capabilities of LLMs \citep{willig2022can, zevceviccausal, qin2019counterfactual,luo2025assistedds,du2024drift}. For example, \citet{kiciman2023causal} demonstrated that LLMs can infer causal relationships only from variable names, outperforming traditional statistical approaches~\citep{peters2017elements}. However, these evaluations focus on scenarios involving commonsense causality. To bridge this gap, \citet{jin2023cladder} introduced synthetic datasets generated from causal graphs, thereby enabling the assessment of LLMs' causal reasoning performance in contexts extending beyond commonsense knowledge.  Additionally, recent works~\citep{long2023causal, zhou2024causalbench} have assessed LLMs' capabilities in data-driven causal inference tasks, focusing on accuracy in estimating causal effects and recovering DAG structures from observational data.

\section{Benchmark Curation}

\subsection{Pitfall Categories and Challenges}

To evaluate the reliability of causal inference performed by LLMs, we introduce the benchmark \textbf{CausalPitfalls} to assess model performance across common statistical pitfalls. Specifically, our benchmark addresses six major categories of causal inference pitfalls, consisting of 15 distinct challenges. Each challenge includes five questions across difficulty levels ranging from ``very easy'' to ``very hard.''
Table~\ref{tab:benchmark_categories} summarizes the categories and their respective challenges:

\begin{table}[h!]
\centering
\caption{CausalPitfalls benchmark categories and challenges}
\label{tab:benchmark_categories}
\small
\renewcommand{\arraystretch}{1.1}
\scalebox{0.95}{
\begin{tabular}{p{0.43\linewidth} p{0.52\linewidth}}
\toprule
\textbf{Confounding biases and spurious associations} & \textbf{Interventions and experimental reasoning} \\
Simpson's paradox \newline
Selection bias (Berkson's paradox) & 
Observational vs experimental reasoning \newline
Causal effect estimation \\
\addlinespace[4pt]
\textbf{Counterfactual reasoning and hypotheticals} & \textbf{Mediation and indirect causal effects} \\
Counterfactual outcome prediction \newline
Causal necessity and sufficiency &
Mediator-outcome confounding \newline
Sequential mediators \newline
Treatment-mediator interaction effects \\
\addlinespace[4pt]
\textbf{Causal discovery and structure learning} & \textbf{Causal generalization and external validity} \\
Cause-effect direction inference \newline
Handling uncertainty in causal structures &
Population shift and transferability \newline
Temporal stability of causal effects \newline
Contextual interaction and moderation effects \newline
Domain shift and transportability of causal knowledge \\
\bottomrule
\end{tabular}
}
\end{table}

Below is a brief overview of the six major categories in the \textbf{CausalPitfalls} benchmark:

\begin{itemize}[leftmargin=*, itemsep=0pt]
    \item \textbf{Confounding biases and spurious associations:} Covers scenarios where misleading correlations arise due to hidden variables or biased conditioning.
    \item \textbf{Interventions and experimental reasoning:} Focuses on distinguishing correlation from causation through randomized experiments or interventional data, and quantifying treatment effects.
    \item \textbf{Counterfactual reasoning and hypotheticals:} Evaluates LLMs' ability to answer    ``what if'' questions by reasoning about alternate outcomes under hypothetical changes.

    \item \textbf{Mediation and indirect causal effects:} Assesses whether models can identify and reason through intermediate causal pathways, including complex mediation structures.
    \item \textbf{Causal discovery and structure learning:} Tests the ability to infer causal directionality or relationships from data without pre-defined causal graphs.
    \item \textbf{Causal generalization and external validity:} Examines whether causal conclusions generalize across different contexts, populations, and environments.
\end{itemize}

Many of these scenarios can also be framed as purely statistical association problems. However, throughout this benchmark, we adopt a causal interpretation grounded in either the Neyman–Rubin potential outcomes framework~\citep{splawa1990application,rubin1974estimating} or Pearl's do-calculus~\citep{pearl2009causality}. This distinction ensures that our tasks target causal reasoning, rather than surface-level correlations.

One illustrative example of these pitfalls is Simpson's paradox~\citep{simpson1951interpretation} (Fig.~\ref{fig:main_fig}a), a commonly misunderstood statistical phenomenon. Simpson's paradox occurs when overall data seem to suggest one conclusion: for example, patients recover better with a particular medical treatment. At first glance, the treatment might appear beneficial. However, once the patients are divided into subgroups based on age, the same data show that the treatment is harmful within each age group. An incorrect analysis would overlook the importance of these subgroups, mistakenly suggesting the treatment is effective. 
Importantly, the stratification variable in Simpson's paradox is assumed to be a confounder, not a collider.

\subsection{Evaluation Protocols and Data}

\paragraph{LLM-based Causal Inference Protocols.} In this paper, we consider two unified protocols to evaluate LLM reliability~(Fig.~\ref{fig:main_fig}b): 
\textbf{(1) Direct Prompting}: LLMs directly answer causal inference questions based on the provided raw data. This approach tests the models' intrinsic capability to perform causal inference without additional computational tools or external support.
\textbf{(2) Code-Assisted Prompting}: LLMs generate executable code to perform statistical analysis relevant to the questions, then interpret the results to answer the questions. This method assesses the LLMs' ability to translate causal reasoning tasks into accurate computational procedures and use analytical results to avoid common pitfalls.

\paragraph{Questions.}
Each challenge includes five versions of the core question, ranging from very easy to very hard~(Table~\ref{tab:questions}). The easier versions give the model more guidance. For example, pointing out the confounder to adjust for or asking directly about Simpson’s paradox. As the difficulty increases, these hints are gradually removed. This setup lets us test whether models can still recognize and handle causal pitfalls when less direction is given (Fig.~\ref{fig:main_fig}b).

\begin{table}[tb]
\centering
\caption{Questions by Difficulty Level for ``Simpson's Paradox''
}
\label{tab:questions}
\small
\scalebox{0.95}{
\renewcommand{\arraystretch}{1.3}
\begin{tabularx}{\linewidth}{>{\raggedright\arraybackslash}p{2cm}X}
\toprule
\textbf{Difficulty} & \textbf{Question} \\
\midrule
Very Easy & Evaluate whether \texttt{\{TREATMENT\}} causally affects \texttt{\{OUTCOME\}}. Adjust for the known confounder (\texttt{\{CONFOUNDER\}}) using stratification or regression. State if Simpson's paradox is present, and provide adjusted rates with 95\% confidence intervals along with a recommendation. \\[4pt]

Easy & Evaluate whether \texttt{\{TREATMENT\}} causally affects \texttt{\{OUTCOME\}}. Consider the potential influence of the confounder (\texttt{\{CONFOUNDER\}}), adjust accordingly, and assess Simpson's paradox. Provide adjusted effect sizes with uncertainty estimates and a recommendation. \\[4pt]

Medium & Evaluate the causal impact of \texttt{\{TREATMENT\}} on \texttt{\{OUTCOME\}}. Account for relevant confounders. Provide adjusted effects with uncertainty measures and justify your recommendation. \\[4pt]

Hard & Assess the causal relationship between \texttt{\{TREATMENT\}} and \texttt{\{OUTCOME\}}, considering potential confounders. \\[4pt]

Very Hard & Evaluate whether \texttt{\{TREATMENT\}} causally affects \texttt{\{OUTCOME\}} without additional hints. \\[4pt]
\bottomrule
\end{tabularx}
}
\end{table}

\paragraph{Datasets.} 
To construct datasets tailored to each challenge, we utilize causal graphs following~\citet{pearl2000models} and~\citet{peters2017elements}. For every statistical pitfall, we select causal graphs that capture its unique complexities and characteristics. Each challenge is accompanied by five distinct datasets, each containing over 500 samples for comprehensive evaluation.
Our simulation approach uses structural causal models based on directed acyclic graphs (DAGs), where each structural equation represents a causal mechanism rather than merely a statistical association. The coefficients in these equations directly encode the causal effects, allowing us to define the ground truth against which inference methods can be evaluated. This approach is mathematically equivalent to simulating potential outcomes under the specified causal structure. The structural equations include both linear and non-linear forms (e.g., non-linear
  link functions and interaction terms), so the evaluation is not limited to purely
  linear relationships.

\subsection{Evaluation Metrics}

To evaluate the reliability of LLMs for causal inference, we developed detailed grading rubrics for each causal pitfall, informed by guidelines from~\citet{sterne2016robins,vandenbrouckel2007strengthening}. Each benchmark challenge includes multiple questions, each assigned points based on how effectively the model addresses the specific pitfall (see Appendix for detailed rubric). The total \textit{score} for a challenge is the sum of points obtained across these questions, and \textit{max\_score} is the maximum achievable score. To enable fair comparisons across challenges, we compute a normalized score:
\begin{equation}\label{eq:normalized_score}
    \text{Normalized Score (\%)} = \frac{\text{score}}{\text{max\_score}} \times 100\%.
\end{equation}

We evaluate LLM responses automatically using an independent GPT-4o model~\citep{achiam2023gpt} to minimize potential biases. To validate the accuracy of this automated evaluation, we additionally engaged three statisticians to manually grade 150 randomly selected responses. We measure consistency between automated and human scores using the \textit{gap} metric:
\[
\text{Gap} = \frac{1}{150}\sum_{i=1}^{150} \frac{|\text{score}_{\text{LLM}}^{(i)} - \text{score}_{\text{human}}^{(i)}|}{s_{\max,i}}\in [0,1],
\]
where \(s_{\max, i}\) is the maximum score of corresponding challenge, and \(\text{score}_{\text{LLM}}^{(i)},~\text{score}_{\text{human}}^{(i)}\in\mathbb{N}^{+}\) are scores from automated and human evaluations, respectively. Here, the gap metric ranges from 0 to 1, where 0 indicates perfect agreement.

Finally, to provide a summary metric, we define \textit{causal reliability} as the average normalized score across all benchmark challenges. This measure captures the overall trustworthiness and reliability of LLMs in statistical causal inference tasks.
\vspace{-0.1cm}

\section{Illustrative Pitfalls in Causal Reasoning} 
\label{sec:two_pitfalls}
\vspace{-0.1cm}

When evaluating causal inference with LLMs, surface-level answers can create an illusion of competence. Statistical causal inference requires grounding conclusions in evidence, checking assumptions, and ruling out alternatives. LLMs, however, may produce confident but flawed outputs that rely on irrelevant cues or statistical artifacts, giving a false sense of robustness.

We illustrate this problem with two failure cases. The first shows how models can base conclusions on superficial semantic cues instead of data, while the second shows how they may misinterpret random variation as causal structure.

\paragraph{Branding Bias: Adversarial Sensitivity to Branding and Semantic Manipulation.}
As an illustrative example, we examined whether LLMs can be misled by superficial cues when drawing causal conclusions. We constructed a synthetic scenario in which beverage consumption affected health outcomes. In the setting, other factors, lifestyle and health awareness, acted as confounders, but the brand label itself had no causal role (Fig.~\ref{fig:branding_bias_dag}).

\begin{figure}[ht!]
\centering
\begin{tikzpicture}[
    scale=0.7,
    transform shape,
    node distance=1.2cm,
    >={Stealth[round]},
    every node/.style={
      draw,
      rounded corners,
      align=center,
      minimum width=3cm,
      minimum height=1cm
    },
    every path/.style={draw, thick}
]

\node (Awareness) {Health Awareness};
\node (Lifestyle) [right=2.8cm of Awareness] {Lifestyle};
\node (Consumption) [below=0.8cm of $(Awareness)!0.5!(Lifestyle)$] {Consumption};
\node (Outcome) [below=0.3cm of Consumption] {Health Outcome};

\draw[->] (Awareness) -- (Consumption);
\draw[->] (Lifestyle) -- (Consumption);
\draw[->] (Awareness.south) |- (Outcome.west);
\draw[->] (Lifestyle.south) |- (Outcome.east);
\draw[->] (Consumption) -- (Outcome);

\end{tikzpicture}
\caption{
Causal DAG illustrating how beverage consumption, health awareness, and lifestyle affect health outcomes. The beverage's brand name (``HealthPlus'' or ``UltraSugar'') does not causally influence outcomes.
}
\label{fig:branding_bias_dag}
\end{figure}
\vspace{-0.2cm}

Despite identical data, simply changing the brand name from a healthy-sounding label (\textit{HealthPlus}) to a harmful-sounding one (\textit{UltraSugar}) induced changes in LLMs' conclusions (Table~\ref{tab:branding_bias_results}). GPT-4o and Gemini-2.0-flash, for example, made a conclusion aligning purely with the brand semantics, rather than the provided data.

The branding bias example shows that LLMs may rely on superficial semantic cues, attributing causal effects to labels even when the underlying data provide evidence \textbf{to the contrary}.

\vspace{-0.2cm}
\begin{table}[ht!]
\caption{Branding bias evaluation. Each row represents a combination of the beverage’s label (“HealthPlus” or “UltraSugar”) and the true effect of the given data (beneficial or harmful). A checkmark (\checkmark) indicates a beneficial conclusion and a cross ($\times$) a harmful conclusion; correct inferences are those that match the true effect.}
\vspace{0.2cm}
\label{tab:branding_bias_results}
\centering
\small
\begin{tabular}{l c c c c}
\toprule
\textbf{Brand Label} & \textbf{True Effect (Data)} & \textbf{GPT-4o} & \textbf{Gemini-2.0-flash} & \textbf{Claude-3.5-sonnet} \\
\midrule
HealthPlus  & \checkmark & \checkmark & \checkmark & \checkmark \\
UltraSugar  & \checkmark & $\times$   & $\times$   & \checkmark \\
HealthPlus  & $\times$     & \checkmark & \checkmark & \checkmark \\
UltraSugar  & $\times$     & $\times$   & $\times$   & \checkmark \\
\bottomrule
\end{tabular}
\end{table}

\paragraph{Spurious Causal Inference from Random Patterns.}  As a second illustrative example, we show how LLMs can mistake random variation in real-world data for genuine causal structure.  Specifically, we evaluated LLMs on real data from a PNAS study~\citep{van2015gender} that analyzed funding success rates across academic disciplines in the Netherlands. When asked if the data reveal gender bias favoring male applicants, all of the tested LLMs drew incorrect conclusions, either attributing the differences in raw percentages directly to gender or considering Simpson’s paradox.

However, a careful statistical analysis as illustrated in the Appendix following \citet{irizarry2019introduction} shows that women are not disproportionately applying to more competitive disciplines, so Simpson’s paradox does not apply. Furthermore, computing log-odds ratios divided by their standard errors across disciplines and adjusting for multiple comparisons reveals no statistical evidence of gender bias within any department. None of the tested LLMs recognized these key insights.

The spurious inference example shows that LLMs may mistake random variation in observational data for genuine causal structure, failing to apply the statistical checks needed to rule out noise.

\section{Results and Analysis}

In this section, we present the key findings from our experiments, examining the causal inference capabilities of ten large language models.  
Our evaluation spans both closed source and open source LLMs, with the complete list of models provided in the Appendix.  
All models were evaluated using two protocols: direct prompting and code assisted prompting.  
The assessment covered six categories of causal inference pitfalls and fifteen challenges, each built from five datasets and five questions of varying difficulty.

\begin{table}[!ht]
\caption{Causal reliability across causal pitfalls, comparing direct and code-assisted prompting. Values represent averages of normalized scores, defined in equation~\eqref{eq:normalized_score}, across five questions per pitfall category; higher scores indicate better performance.}
\vspace{0.2cm}
\label{tab:overall_performance}
\centering
\small
\setlength{\tabcolsep}{4pt}
\scalebox{1}{
\begin{tabular}{lccccccc}
\toprule
\textbf{LLM (Direct Prompting)} & Conf & Interv & Counter & Med & Disc & Ext & Average \\
\midrule
Gemma2-9b & 14.00 & 30.72 & 8.00 & 15.89 & 9.49 & 4.05 & 13.69 \\
Llama3.1-8b & 18.46 & 34.78 & 10.86 & 29.33 & 7.71 & 6.05 & 17.86 \\
Llama3.1-70b & 17.60 & 35.92 & 10.29 & 33.67 & 11.31 & 9.92 & 19.78 \\
Mistral-7b & 17.31 & 29.83 & 5.71 & 19.17 & 8.40 & 6.18 & 14.43 \\
Mixtral-8x22b & 16.40 & 32.08 & 8.00 & 27.50 & 9.20 & 8.75 & 16.99 \\
Claude-3.5-sonnet & 18.74 & 47.60 & 12.00 & 40.50 & 18.63 & 19.84 & 26.22 \\
Gemini-2.0-flash & 20.06 & 37.56 & 13.43 & 46.67 & 13.48 & 14.93 & 24.36 \\
Deepseek-chat & 25.89 & \textbf{52.42} & 12.86 & 53.83 & 20.83 & 28.74 & 32.43 \\
GPT-4.1 & 17.26 & 33.57 & 6.57 & 53.27 & 16.43 & 24.31 & 25.24 \\
GPT-o4-mini & \textbf{41.43} & 45.21 & \textbf{18.57} & \textbf{57.67} & \textbf{36.97} & \textbf{44.48} & \textbf{40.72} \\
\bottomrule
\\[8pt]
\toprule
\textbf{LLM (Code-Assisted Prompting)} & Conf & Interv & Counter & Med & Disc & Ext & Average \\
\midrule
Gemma2-9b & 7.89 & 18.50 & 4.86 & 23.04 & 6.63 & 18.50 & 13.24 \\
Llama3.1-8b & 11.93 & 15.99 & 7.67 & 18.06 & 6.23 & 17.29 & 12.86 \\
Llama3.1-70b & 23.54 & 22.83 & 8.21 & 25.17 & 14.41 & 25.17 & 19.89 \\
Mistral-7b & 4.68 & 13.15 & 1.43 & 11.11 & 6.43 & 9.09 & 7.65 \\
Mixtral-8x22b & 22.50 & 30.53 & 5.44 & 26.49 & 14.29 & 24.31 & 20.59 \\
Claude-3.5-sonnet & 32.91 & 47.33 & 11.71 & 41.73 & 16.12 & 26.47 & 29.38 \\
Gemini-2.0-flash & 37.20 & 42.96 & 14.29 & 42.17 & 16.19 & 37.98 & 31.80 \\
Deepseek-chat & 38.63 & 48.70 & 10.86 & 47.13 & 25.79 & 45.63 & 36.12 \\
GPT-4.1 & 47.14 & 42.65 & 12.29 & 49.40 & 23.87 & 48.58 & 37.32 \\
GPT-o4-mini & \textbf{62.00} & \textbf{51.86} & \textbf{16.96} & \textbf{50.00} & \textbf{26.67} & \textbf{50.71} & \textbf{43.03} \\
\bottomrule
\end{tabular}
}
\vspace{2pt}
\begin{flushleft}
{\footnotesize\textit{
Conf: Confounding biases and spurious associations; Interv: Interventions and experimental reasoning; Counter: Counterfactual reasoning and hypotheticals; Med: Mediation and indirect causal effects; Disc: Causal discovery and structure learning; Ext: Causal generalization and external validity.
}}
\end{flushleft}
\end{table}
\vspace{-0.2cm}
\paragraph{Overall Performance.}
Across models, \textbf{GPT-o4-mini} demonstrated the highest overall reliability, achieving an average causal reliability of $40.72\%$ under direct prompting and $43.03\%$ under code-assisted prompting (Table~\ref{tab:overall_performance}).  
Deepseek-chat, although lower in aggregate (average 32.43\% under direct prompting and 36.12\% under code-assisted prompting), obtained the strongest performance in \textit{interventions and experimental reasoning} (52.42\% under direct prompting and 48.70\% under code-assisted prompting).
Taken together, these results indicate that optimized mid-scale models can, in certain contexts, outperform larger frontier systems on causal reasoning tasks.

\paragraph{Benefits of Code-Assisted Prompting.}
Although code-assisted prompting improved several strong models, its benefits were not universal.  
For example, GPT-o4-mini and Gemini-2.0-flash gained from code execution, increasing their averages from $40.72\%$ to $43.03\%$ and $24.36\%$ to $31.80\%$, respectively.  
GPT-4.1 also improved significantly ($25.24\% \rightarrow 37.32\%$).  Under code-assisted prompting, models first convert raw data into summary statistics  
  through generated code, then reason over those statistics. This separates low-level    
  data parsing from causal reasoning, which partly explains why stronger models benefit  
  more: they produce correct analysis code and reason on clean numerical outputs rather 
  than raw tables.
In contrast, some small open-source LLMs showed little benefit: Llama3.1-8B decreased from $17.86\%$ to $12.86\%$, Gemma2-9B remained unchanged ($13.69\%$ to $13.24\%$), and Mistral-7B even dropped from $14.43\%$ to $7.65\%$.  
These results indicate that computational results increase the strengths of already-capable LLMs, but does not provide a uniform advantage across the LLMs.  Models with high code-error rates (e.g., Mistral-7B, Llama-8B) have worse scores under
  code-assisted prompting. Allowing one debugging attempt can bring their performance back
  to direct-prompting levels. We discuss this further in
  Appendix~\ref{sec:debugging_protocol}.

\paragraph{Impact of Difficulty Levels.}
Causal reliability consistently decreases as the question becomes harder.  
As shown in Table~\ref{tab:difficulty_performance}, all models performed best on very easy and easy questions, with average scores declining as task difficulty increased, where the question contains fewer hints~(see Table~\ref{tab:questions}.  
For example, GPT-o4-mini achieved $60.72\%$ under direct prompting and $56.73\%$ under code-assisted prompting on very easy items, but dropped to $17.75\%$ and $32.84\%$, respectively, on very hard questions.  
Deepseek-chat and Gemini-2.0-flash showed similar trends, maintaining moderate performance on medium and hard levels but falling quickly on the \textit{very hard} questions.  
These results indicate that code assistance is particularly beneficial for challenging tasks, helping stronger models recover some performance at higher difficulty levels.

\begin{table}[ht]
\caption{Causal reliability by difficulty levels of questions, comparing direct and code-assisted prompting. Values represent averages of normalized score, defined in equation~\eqref{eq:normalized_score}, across 15 challenges; higher scores indicate better performance.}
\label{tab:difficulty_performance}
\centering
\small
\setlength{\tabcolsep}{4pt}
\scalebox{1}{
\begin{tabular}{lccccc}
\toprule
\textbf{LLM (Direct Prompting)} & Very Easy & Easy & Medium & Hard & Very Hard \\
\midrule
Gemma2-9b & 20.50 & 15.64 & 14.22 & 6.72 & 5.46 \\
Llama3.1-8b & 27.99 & 20.85 & 19.90 & 10.71 & 5.82 \\
Llama3.1-70b & 31.04 & 27.38 & 22.04 & 11.29 & 5.23 \\
Mistral-7b & 20.04 & 18.88 & 18.02 & 7.49 & 3.80 \\
Mixtral-8x22b & 27.80 & 21.20 & 20.31 & 8.41 & 5.23 \\
Claude-3.5-sonnet & 40.27 & 36.63 & 29.69 & 13.42 & 8.30 \\
Gemini-2.0-flash & 37.83 & 33.59 & 27.59 & 15.72 & 8.22 \\
Deepseek-chat & 46.37 & 38.92 & 37.92 & 28.79 & 16.36 \\
GPT-4.1 & 38.01 & 33.94 & 33.59 & 17.31 & 7.67 \\
GPT-o4-mini & \textbf{60.72} & \textbf{54.05} & \textbf{48.13} & \textbf{30.72} & \textbf{17.75} \\
\bottomrule
\\[8pt]
\toprule
\textbf{LLM (Code-Assisted Prompting)} & Very Easy & Easy & Medium & Hard & Very Hard \\
\midrule
Gemma2-9b & 19.09 & 18.35 & 16.43 & 10.56 & 7.94 \\
Llama3.1-8b & 26.04 & 14.41 & 12.77 & 8.84 & 7.90 \\
Llama3.1-70b & 32.21 & 27.88 & 24.77 & 14.08 & 7.58 \\
Mistral-7b & 9.44 & 12.03 & 9.59 & 6.63 & 2.73 \\
Mixtral-8x22b & 33.14 & 29.77 & 25.64 & 14.59 & 7.77 \\
Claude-3.5-sonnet & 39.70 & 35.06 & 33.55 & 20.92 & 18.34 \\
Gemini-2.0-flash & 46.01 & 39.83 & 32.17 & 28.00 & 21.67 \\
Deepseek-chat & 55.56 & 48.23 & 43.69 & 28.46 & 16.50 \\
GPT-4.1 & \textbf{57.80 }& 46.49 & 46.48 & 24.33 & 19.65 \\
GPT-o4-mini & 56.73 & \textbf{49.66} & \textbf{49.37} & \textbf{34.31} & \textbf{32.84} \\
\bottomrule
\end{tabular}
}
\end{table}

\paragraph{Persistent Reliability Gaps.}
Despite variation across models and prompting strategies, significant reliability gaps remain in all settings.  
Among the tested LLMs,  GPT-o4-mini achieved the highest causal reliability score $43.03\%$ on average under code-assisted prompting, while most other models remained well below this level.  
Performance on difficult causal reasoning challenges was especially limited, with \textit{very hard } questions rarely exceeding $30\%$ even for the best LLMs.  
Taken together, the results suggest that existing LLMs, when used without finetuning or specialized architectures, remain unreliable to apply in high-stakes causal inference.

\paragraph{Human-LLM Grading Alignment.}
To validate the fidelity of our automated GPT-4o scoring against expert judgments, we conducted a human validation study on a stratified sample of 150 model responses, with equal representation across the six pitfall categories and five difficulty levels (see Appendix). Three PhD students in statistics independently graded each response using our detailed rubrics. We then compared the resulting human scores with those produced by GPT-4o via the gap metric, which yielded a mean value of 0.11. This close agreement confirms that our GPT-4o evaluator reliably mirrors expert assessments, justifying its use for large-scale, reproducible evaluation of LLM performance in causal inference without the need for extensive human oversight.

\begin{figure}[ht]
    \centering
    \includegraphics[width=\linewidth]{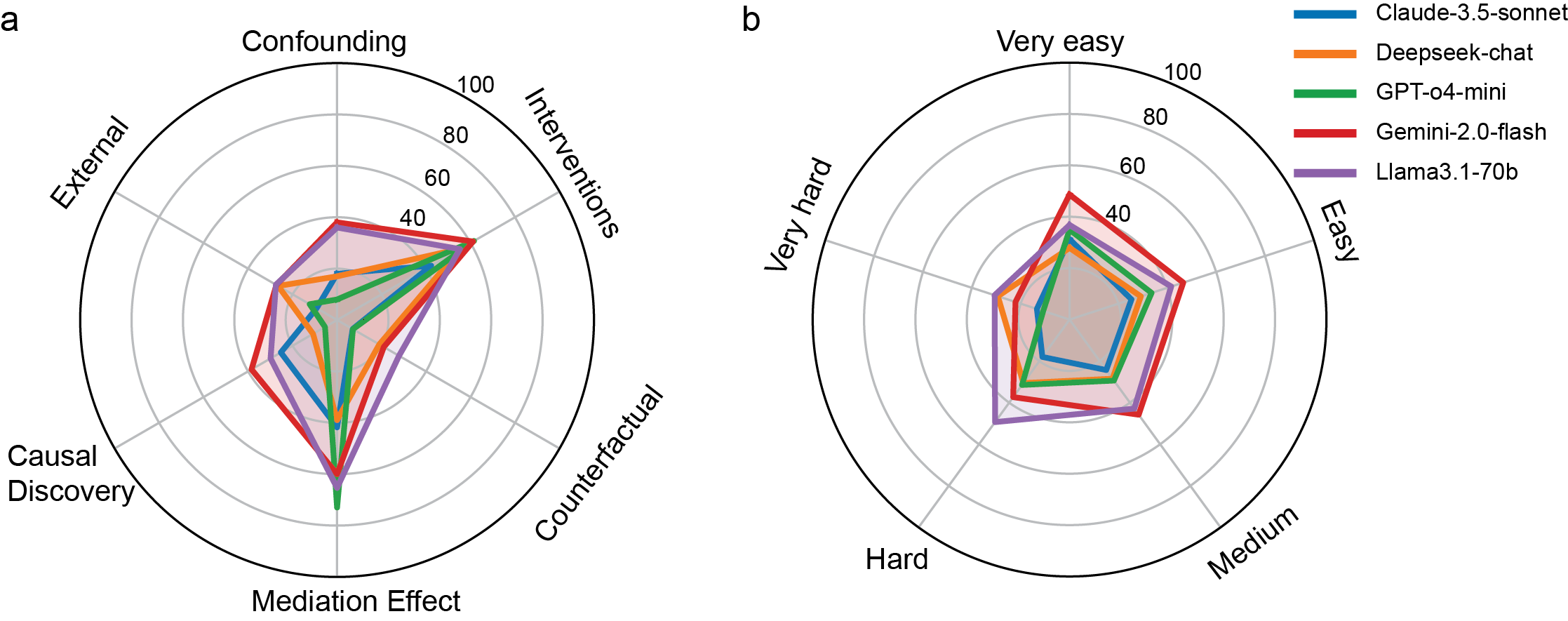}
    \caption{\textbf{Code execution failure rates (\%) in code-assisted prompting protocol across causal inference challenges and question difficulty.}  
    Failure rate is defined as the percentage of code-generation attempts that either raise execution errors or produce invalid analytical outputs, computed only for the code-assisted prompting protocol.  
    (a) Average failure rate for each of the six causal-inference pitfall categories.  
    (b) Average failure rate by question difficulty level, increasing from very easy through very hard tasks.
    }
    \label{fig:error_rate_radar}
\end{figure}

\paragraph{Code-Assisted Execution Errors.}
As shown in Figure~\ref{fig:error_rate_radar}, code‐execution failures peak in the ``mediation effects'' and ``interventions and experimental reasoning'' categories, where implementing correct stratification and transportability routines is most demanding. Interestingly, ``very easy'' questions produce the highest failure rates, whereas ``very hard'' questions yield lower rates.   

\section{Conclusion}

We introduced \textbf{CausalPitfalls}, a benchmark designed to rigorously evaluate the reliability of LLMs in performing statistical causal inference. Unlike existing benchmarks that focus primarily on accuracy, our benchmark reveals how LLMs can produce confident yet flawed conclusions by falling into classical statistical pitfalls. 
Our results indicate substantial gaps in reliability across all models and settings. Even state-of-the-art models exhibit systematic vulnerabilities to confounding, semantic bias, and difficulties in generalizing causal knowledge across contexts. These findings highlight an urgent need for targeted interventions to improve LLMs' trustworthiness in scientific and policy domains.
Future directions include expanding the benchmark to cover more nuanced forms of causal reasoning, such as instrumental variable analysis, latent confounding, and policy evaluation. Additionally, we envision CausalPitfalls as a platform to guide training or fine-tuning strategies that aim to instill causal robustness in LLMs. 

More detailed descriptions of our benchmark pitfall categories, challenges, and implementation details are included in the \textbf{Appendix} inside the supplementary material.

\subsubsection*{Acknowledgment} 

The work was supported in part by the National Science Foundation CAREER Program under grant number 2338506.

\subsubsection*{Ethics statement}

All authors have read and adhered to the ICLR Code of Ethics.  

\subsubsection*{Reproducibility statement}

We have made every effort to ensure the reproducibility of our results. Detailed descriptions of datasets, preprocessing steps, and experimental settings are provided in the main text and appendix. The CausalPitfalls    
  benchmark, including all datasets, questions, grading rubrics, and 
  evaluation scripts, is publicly available at 
  \url{https://github.com/dudududuu/CausalPitfalls}.

\bibliography{bib}
\bibliographystyle{iclr2026_conference}

\clearpage
\appendix

\section{The Use of Large Language Models (LLMs)}

The authors acknowledge the use of ChatGPT to check for potential typographical and grammatical errors in the manuscript.

\section{Benchmarking key challenges in reliable causal inference for LLMs}

\subsection{Confounding Biases and Spurious Associations}

\paragraph{Simpson’s Paradox.}

Simpson’s paradox is a causal pitfall where an observed association between two variables reverses or disappears when analyzed within subgroups defined by a confounding variable. This paradox typically arises when aggregated data conceal important subgroup-specific relationships, making an apparent correlation misleading or even contradictory.

\textit{Motivation and Relevance:}  
Addressing Simpson’s paradox is crucial because misinterpretations of aggregated relationships can lead to serious real-world consequence. LLMs are especially susceptible to this pitfall due to their tendency to overly rely on surface-level correlations and aggregated statistics, while neglecting underlying subgroup structures. Without specifically considering these subgroup, LLMs risk confidently delivering inaccurate causal conclusions.

\textit{Data Generation:}  
We design five datasets to illustrate Simpson’s paradox, and each dataset contains three binary variables representing a realistic medical scenario. Here's an example of data generation setting:

\begin{itemize}[leftmargin=*, itemsep=1pt, topsep=3pt]
    \item Age Group (Confounder): Represents subgroup differences that might influence treatment assignment and outcomes (\textit{Young} or \textit{Old}).
    \item Drug Treatment (Treatment): Indicates whether a patient received a specific medication (\textit{Drug given} or \textit{No drug}).
    \item Recovery Status (Outcome): Represents patient recovery (\textit{Recovered} or \textit{Not recovered}).
\end{itemize}

The causal structure underlying the simulation is shown by the DAG below:

\begin{center}
\begin{tikzpicture}[node distance=18mm, every node/.style={
    draw,
    rectangle,
    rounded corners,
    align=center,
    inner sep=4pt
  }]
  \node (Age)       {Age Group};
  \node[right=of Age] (Drug)      {Drug Treatment};
  \node[below=of Drug] (Recovery) {Recovery Status};

  \draw[->] (Age) -- (Drug);
  \draw[->] (Age) -- (Recovery);
  \draw[->] (Drug) -- (Recovery);
\end{tikzpicture}
\end{center}

In this DAG, Age Group affects both the likelihood of receiving the drug treatment and the probability of recovery. Thus it confounds the relationship between treatment and recovery status. The generated datasets illustrate Simpson’s paradox: when analyzed at an aggregate level, the drug appears beneficial, but within each age subgroup, the drug is actually harmful.

\textit{Evaluation Rubric:}  
Models are evaluated based on their ability to recognize and correct for Simpson’s paradox. Full credit (total score) is awarded if the LLM acknowledges the confounder, correctly identifies subgroup-specific relationships, and accurately states that the aggregate-level relationship is misleading. Partial credit is given if the model recognizes the discrepancy but fails to identify or correctly adjust for the confounder. No credit is awarded if the model relies only on aggregate-level associations without mentioning subgroup analysis or confounding.

\paragraph{Berkson’s Paradox}

Berkson’s paradox is a causal pitfall arising from conditioning on a collider variable, thereby inducing a spurious association between two variables that are actually independent. Typically, this paradox appears when the analysis is restricted to a subgroup selected based on a variable influenced simultaneously by two independent factors, creating an artificial correlation.

\textit{Motivation and Relevance:}  
Recognizing Berkson’s paradox is essential, as it can lead to severely misguided conclusions. LLMs are particularly prone to this pitfall due to their reliance on readily accessible data summaries and correlations without carefully assessing selection mechanisms or collider structures. Without  addressing collider-induced biases, LLMs may confidently report spurious relationships as meaningful causal insights.

\textit{Data Generation:}  
We design five datasets to illustrate Berkson’s paradox across realistic scenarios. Each dataset involves three binary variables, with one serving as a collider that induces selection bias. An example dataset represents a hospital-based scenario:

\begin{itemize}[leftmargin=*, itemsep=1pt, topsep=3pt]
    \item Disease A: Indicates whether a patient has Disease A (\textit{Present} or \textit{Absent}).
    \item Disease B: Indicates whether a patient has Disease B (\textit{Present} or \textit{Absent}).
    \item Hospitalization (Collider): Represents whether the patient is hospitalized (\textit{Yes} or \textit{No}), influenced by both diseases.
\end{itemize}

The causal structure underlying Berkson’s paradox is depicted by the DAG below:

\begin{center}
\begin{tikzpicture}[node distance=18mm, every node/.style={
    draw,
    rectangle,
    rounded corners,
    align=center,
    inner sep=4pt
  }]
  \node (DiseaseA) {Disease A};
  \node[right=of DiseaseA] (DiseaseB) {Disease B};
  \node[below right=12mm and 0mm of DiseaseA] (Hospital) {Hospitalization};

  \draw[->] (DiseaseA) -- (Hospital);
  \draw[->] (DiseaseB) -- (Hospital);
\end{tikzpicture}
\end{center}

In this DAG, the variable \textit{Hospitalization} is a collider, as it is influenced by Disease A and Disease B. When analyses are conditioned upon or restricted to hospitalized patients, it artificially creates an association between these two otherwise independent diseases. Thus, Berkson’s paradox emerges  through these generated datasets, underscoring the critical importance of avoiding inappropriate conditioning or properly correcting for selection biases.

\textit{Evaluation Rubric:}  
Models are evaluated on their ability to recognize and appropriately correct for Berkson’s paradox. Full credit is awarded if the LLM  proposes the correct collider structure, states the independence assumption between the variables, proposes to use a valid bias-correction method (such as inverse probability weighting), and suggests no true association after correcting for collider bias. Partial credit is awarded if the model identifies collider bias but fails to  state assumptions, proposes the correct correction approach, or  suggests independence. No credit is given if the model neglects the collider issue altogether and draws conclusions based only on na{\"\i}ve associations without appropriate adjustment.

\subsection{Interventions and Experimental Reasoning}

\paragraph{Observational vs Experimental Reasoning}

Observational reasoning refers to inferring causal relationships only from non-experimental, observational data, where the treatment is not randomly assigned. In contrast, experimental reasoning involves randomized controlled trials (RCTs), where treatment assignment is randomized, ensuring that confounding variables are evenly distributed across groups. The causal pitfall emerges when analysts interpret observational associations as causal effects without adequately adjusting for confounding variables, potentially resulting in incorrect causal conclusions.

\textit{Motivation and Relevance:}  
Distinguishing between observational and experimental reasoning is crucial, as erroneous interpretations of observational data can significantly affect policy decisions, medical recommendations, and economic strategies. LLMs are particularly vulnerable to this pitfall due to their propensity to accept apparent observational correlations at face value, often neglecting confounders or the necessity of proper statistical adjustments. Without  instruction or analytical rigor, LLMs may deliver misguided causal interpretations, amplifying the risk of poor decision-making in sensitive real-world applications.

\textit{Data Generation:}  
We design five datasets to illustrate the differences between observational and experimental reasoning across realistic scenarios. Each dataset includes a binary treatment and a binary outcome, along with several confounding variables. Here is an illustrative example scenario focusing on sleep quality:

\begin{itemize}[leftmargin=*, itemsep=1pt, topsep=3pt]
    \item Supplement Use (Treatment): Indicates whether participants take a sleep-improving supplement (\textit{Yes} or \textit{No}).
    \item Sleep Quality (Outcome): Represents participants' quality of sleep (\textit{High quality} or \textit{Low quality}).
    \item True Confounders: Regular exercise habits (\textit{Exercise regularly}) and older age (\textit{Age old}), each affecting both supplement use and sleep quality.
    \item Irrelevant Variables (Non-confounders): Factors such as coffee drinking, watching TV late, and income, included to test the model’s ability to discern relevant confounding from irrelevant variables.
\end{itemize}

The causal structure illustrating observational confounding is depicted by the DAG:

\begin{center}
\begin{tikzpicture}[node distance=18mm, every node/.style={
    draw,
    rectangle,
    rounded corners,
    align=center,
    inner sep=4pt
  }]
  \node (Confounders) {Exercise, Age};
  \node[right=of Confounders] (Treatment) {Supplement Use};
  \node[right=of Treatment] (Outcome) {Sleep Quality};

  \draw[->] (Confounders) -- (Treatment);
  \draw[->] (Confounders) to[bend right=15] (Outcome);
  \draw[->] (Treatment) -- (Outcome);
\end{tikzpicture}
\end{center}

In this DAG, confounders (Exercise and Age) simultaneously affect both the treatment (Supplement Use) and the outcome (Sleep Quality), introducing bias into na{\"\i}ve observational analyses. Proper adjustment for these confounders is essential to approximate the true causal relationship.

\textit{Evaluation Rubric:}  
Models are evaluated based on their capability to distinguish observational correlations from true causal effects. Full credit is awarded if the LLM correctly identifies true confounders, justifies their relevance, applies an appropriate statistical adjustment method (such as stratification or regression), and  distinguishes the corrected causal conclusion from na{\"\i}ve observational results. Partial credit is awarded if the model recognizes the importance of confounding but fails to fully justify confounders or properly apply adjustments. No credit is given if the model relies only on observational correlations without addressing confounding  or providing adjusted causal interpretations.

\paragraph{Causal Effect Estimation}

Causal effect estimation involves quantifying the precise magnitude of the effect that an intervention or treatment has on a particular outcome. Unlike qualitative assessments of causal relationships, effect estimation  focuses on numerical measures, such as the average treatment effect (ATE). The primary challenge arises from properly accounting for confounding factors, distinguishing true causal effects from mere correlations or conditional probabilities.

\textit{Motivation and Relevance:}  
Accurate estimation of causal effects is critical for informed decision-making in domains such as medicine, public policy, marketing, and technology. Misestimating these effects can lead to misguided interventions, inefficient resource allocation, or even unintended harm. LLMs often encounter difficulties with this task due to their inclination to rely on observational correlations without properly adjusting for confounders. Consequently, they risk confusing correlation or conditional probabilities with actual causal effects, potentially delivering flawed or misleading recommendations.

\textit{Data Generation:}  
We design five datasets to evaluate the accuracy and robustness of causal effect estimation methods across realistic scenarios. Each dataset involves a  defined binary treatment, binary outcome, multiple relevant confounders, and irrelevant features to test the ability of the model to distinguish between pertinent and non-pertinent information. An illustrative dataset focuses on patient recovery after secondary medical treatment:

\begin{itemize}[leftmargin=*, itemsep=1pt, topsep=3pt]
    \item Secondary Treatment (Treatment): Indicates whether patients received a second-round medical treatment (\textit{Received} or \textit{Not received}).
    \item Recovery (Outcome): Reflects patient recovery status (\textit{Recovered} or \textit{Not recovered}).
    \item True Confounders: Factors such as patient age, initial illness severity, follow-up severity, and initial treatment, each influencing both the assignment of the secondary treatment and patient recovery.
    \item Irrelevant Features: Variables like socioeconomic status and random noise, intended to evaluate whether the model appropriately excludes irrelevant information.
\end{itemize}

The causal structure underlying these scenarios is represented by the DAG below:

\begin{center}
\begin{tikzpicture}[node distance=18mm, every node/.style={
    draw,
    rectangle,
    rounded corners,
    align=center,
    inner sep=4pt
  }]
  \node (Confounders) {Age, Severity,\\ Initial Treatment};
  \node[right=of Confounders] (Treatment) {Secondary Treatment};
  \node[right=of Treatment] (Outcome) {Recovery};

  \draw[->] (Confounders) -- (Treatment);
  \draw[->] (Confounders) to[bend right=15] (Outcome);
  \draw[->] (Treatment) -- (Outcome);
\end{tikzpicture}
\end{center}

This DAG  shows how confounding variables (Age, Severity, and Initial Treatment) simultaneously affect both the assignment of secondary treatment and the likelihood of recovery. Proper causal estimation requires  statistical adjustments, such as inverse probability weighting (IPW), to isolate the true effect of the treatment from these confounding influences.

\textit{Evaluation Rubric:}  
Models are evaluated based on their ability to accurately estimate causal effects and rigorously justify their methodological decisions. Full credit is awarded if the LLM  defines the target causal estimand (such as ATE), correctly identifies and adjusts for true confounders,  excludes irrelevant features, employs suitable causal inference methods (e.g., inverse probability weighting), provides numerical estimates within an acceptable tolerance range, and appropriately quantifies statistical uncertainty. Additionally, full-scoring responses perform at least one diagnostic check (e.g., balance or overlap assessments) and  discuss methodological limitations. Partial credit is awarded if some, but not all, of these criteria are adequately addressed. No credit is given for relying only on unadjusted correlations without proper causal reasoning or methodological justification.

\subsection{Counterfactual Reasoning and Hypotheticals}
\paragraph{Counterfactual Outcome Prediction}

Counterfactual outcome prediction involves reasoning about hypothetical scenarios, specifically asking what would have happened if past events or treatments had occurred differently. This type of reasoning goes beyond mere correlation, requiring the careful consideration of alternative scenarios that contradict observed reality, while maintaining logical consistency with established causal relationships.

\textit{Motivation and Relevance:}  
Counterfactual reasoning is essential for robust explanations, policy analysis, and informed decision-making across various disciplines, including medicine, economics, education, and environmental policy. Incorrect counterfactual predictions can lead to flawed policy recommendations and misguided interventions. LLMs often struggle with this task, as it requires moving beyond observed data toward hypothetical worlds that differ from reality. LLMs frequently either adhere too closely to observed facts or generate predictions that violate known causal structures, thereby compromising the reliability of their predictions.

\textit{Data Generation:}  
We design five datasets for counterfactual outcome prediction tasks across diverse real-world contexts. Each dataset includes  defined treatment, outcome, confounder, and downstream variables, enabling  manipulation and assessment of counterfactual scenarios. One illustrative example is a clinical scenario involving drug dosage:

\begin{itemize}[leftmargin=*, itemsep=1pt, topsep=3pt]
    \item Drug Dose (Treatment): Dosage levels administered to patients.
    \item Blood Concentration (Outcome): The resulting concentration level in patient blood following dosage.
    \item Baseline Health (Confounder): Initial health condition influencing both drug dosage and blood concentration.
    \item Therapeutic Score (Downstream Variable): A health outcome measure influenced by blood concentration but not directly adjustable in the counterfactual scenario.
\end{itemize}

The causal structure governing these scenarios is represented by the DAG:

\begin{center}
\begin{tikzpicture}[node distance=12mm, every node/.style={
    draw,
    rectangle,
    rounded corners,
    align=center,
    inner sep=4pt
  }]
  \node (Confounder) {Baseline Health};
  \node[right=of Confounder] (Treatment) {Drug Dose};
  \node[right=of Treatment] (Outcome) {Blood Concentration};
  \node[right=of Outcome] (Downstream) {Therapeutic Score};

  \draw[->] (Confounder) -- (Treatment);
  \draw[->] (Confounder) to[bend right=15] (Outcome);
  \draw[->] (Treatment) -- (Outcome);
  \draw[->] (Outcome) -- (Downstream);
\end{tikzpicture}
\end{center}

In this DAG, Baseline Health is a confounder affecting both the treatment (Drug Dose) and the outcome (Blood Concentration). The Therapeutic Score is a downstream variable, influenced only by the outcome and thus not to be controlled when evaluating counterfactuals. This challenge  asks for predictions under hypothetical alterations of the treatment variable, carefully ensuring consistency with the causal framework.

\textit{Evaluation Rubric:}  
Models are evaluated based on their capability to accurately generate and reason about counterfactual outcomes. Full credit is awarded if the LLM  identifies treatment, outcome, confounder, and downstream variables, proposes an appropriate causal DAG,  acknowledges underlying assumptions (e.g., absence of hidden confounding), applies a valid method (such as regression adjustment) to estimate the counterfactual outcome, and provides numeric predictions that align closely with known counterfactual values. Partial credit is granted if the model correctly addresses some but not all aspects, such as identifying variables and providing accurate reasoning but failing to provide numerical accuracy or  state assumptions. No credit is given if the model ignores causal structure, confounding, or essential assumptions, providing counterfactual predictions based only on observed correlations without causal reasoning.

\paragraph{Causal Necessity and Sufficiency}

Evaluating causal necessity and sufficiency involves determining whether a specific factor is required (necessary) or alone capable (sufficient) to produce an outcome. Necessity examines if the outcome would fail to occur in the absence of the cause, whereas sufficiency assesses if the presence of the cause alone invariably leads to the outcome. This analysis is particularly crucial in scenarios with multiple, potentially redundant or overlapping causal factors.

\textit{Motivation and Relevance:}  
Correctly assessing causal necessity and sufficiency is fundamental for rigorous explanation, accountability, and policy decisions. Misclassification can lead to incorrect attribution, misguided interventions, or failure to identify effective alternative measures. LLMs commonly encounter difficulties with this sophisticated causal reasoning; they often erroneously assume necessity in the presence of alternative sufficient causes or incorrectly attribute sufficiency without thorough evaluation. Expert causal reasoning involves considering counterfactual scenarios --- imagining the outcome under conditions where a causal factor is removed or is the sole influencing factor.

\textit{Data Generation:}  
We simulate five datasets  designed to test the understanding of causal necessity and sufficiency across diverse applied contexts. Each scenario includes a focal causal factor (\(X\)), outcome variable (\(Y\)), and additional contextual factors. An illustrative example involves material stress in engineering structures:

\begin{itemize}[leftmargin=*, itemsep=1pt, topsep=3pt]
    \item Material Stress (Focal Cause): Evaluates whether elevated material stress is necessary or sufficient for structural failure.
    \item Failure Probability (Outcome): Indicates whether structural failure occurs beyond a specific threshold.
    \item Contextual Factors: Load pressure and vibration frequency, serving as potential alternative or complementary causal factors influencing failure.
\end{itemize}

The causal structure underlying necessity and sufficiency is illustrated by the DAG:

\begin{center}
\begin{tikzpicture}[node distance=18mm, every node/.style={
    draw,
    rectangle,
    rounded corners,
    align=center,
    inner sep=4pt
  }]
  \node (Factor1) {Load Pressure};
  \node[right=of Factor1] (CauseX) {Material Stress};
  \node[right=of CauseX] (Factor2) {Vibration Frequency};
  \node[below=of CauseX] (Outcome) {Structural Failure};

  \draw[->] (Factor1) -- (Outcome);
  \draw[->] (CauseX) -- (Outcome);
  \draw[->] (Factor2) -- (Outcome);
\end{tikzpicture}
\end{center}

In this DAG, the focal causal factor (Material Stress) is evaluated alongside alternative or complementary causes (Load Pressure, Vibration Frequency) to determine if it is necessary or sufficient for structural failure. Proper analysis requires reasoning about interventions and counterfactual scenarios, comparing outcomes under varying causal conditions.

\textit{Evaluation Rubric:}  
Models are evaluated based on their capability to rigorously define and correctly classify causal necessity and sufficiency. Full credit is awarded if the LLM  defines global necessity and sufficiency using accurate counterfactual reasoning,  mentions critical thresholds, thoroughly analyzes outcomes under varied interventions, describes potential outcomes , and correctly classifies the causal factor based on the provided scenarios. Partial credit is awarded for responses that partially fulfill these criteria—such as accurately defining concepts and analyzing some interventions but not comprehensively covering all required elements. No credit is given if the model fails to apply proper counterfactual logic, ignores alternative causes, or incorrectly classifies necessity and sufficiency without justification.

\subsection{Mediation and Indirect Causal Effects}

Mediation analysis focuses on understanding how a treatment influences an outcome through intermediate variables known as mediators. Identifying these indirect pathways accurately is essential, as interventions typically propagate through multiple, intertwined channels. However, precise estimation of mediated effects raises methodological challenges. 

\paragraph{Mediator–Outcome Confounding}

Mediator–outcome confounding arises when a third variable influences both the mediator (\(M\)) and the outcome (\(Y\)), biasing the estimated indirect (mediation) and direct effects of a treatment (\(T\)). In the canonical structure \(T \!\rightarrow\! M \!\rightarrow\! Y\), a confounder (\(C\)) that affects both \(M\) and \(Y\) distorts mediation estimates unless it is properly controlled. The task therefore requires distinguishing true mediation pathways from correlations induced by common causes of the mediator and the outcome.

\textit{Motivation and Relevance:}  
Failure to address mediator–outcome confounding can lead to severely biased causal interpretations and misguided policies. For instance, when evaluating whether physical activity mediates the effect of an educational program on cognitive performance, socioeconomic status may independently influence both activity levels and test scores. LLMs frequently overlook this bias: they may attribute the entire treatment effect to the mediating variable or misestimate indirect and direct effects because they ignore hidden pathways. Correct reasoning demands recognising potential confounders of \(M\) and \(Y\) and applying specialised methods, such as sequential g-estimation, inverse‐probability–weighted mediation analysis, or parametric g-formulae, to obtain unbiased mediation effects.

\textit{Data Generation:}  
We construct five datasets that embed mediator–outcome confounding in realistic settings. A representative scenario examines an educational intervention:

\begin{itemize}[leftmargin=*, itemsep=1pt, topsep=3pt]
  \item Treatment (\(T\)): Receipt of a study-skills intervention (\textit{Yes} / \textit{No}).
  \item Mediator (\(M\)): Weekly physical-activity hours.
  \item Outcome (\(Y\)): Post-intervention cognitive test score.
  \item Confounder (\(C\)): Socioeconomic status (SES), influencing both activity and test performance.
\end{itemize}

The causal structure is illustrated by the DAG below:

\begin{center}
\begin{tikzpicture}[node distance=15mm, every node/.style={
    draw,
    rectangle,
    rounded corners,
    align=center,
    inner sep=4pt
  }]
  \node (T) {Study-skills Intervention};
  \node[right=of T] (M) {Physical-activity Hours};
  \node[right=of M] (Y) {Cognitive Test Score};
  \node[above=of M] (C) {Socioeconomic Status};

  \draw[->] (T) -- (M);
  \draw[->] (M) -- (Y);
  \draw[->] (T) to[bend right=15] (Y);
  \draw[->] (C) -- (M);
  \draw[->] (C) -- (Y);
\end{tikzpicture}
\end{center}

\textit{Evaluation Rubric:}  
Models are evaluated on their ability to detect and correct for mediator–outcome confounding. Full credit is awarded when the response  identifies the confounder that influences both the mediator and the outcome, employs and justifies a valid adjustment method (for example, sequential \(g\)-estimation or inverse-probability weighting), provides numerical estimates of the natural direct and indirect effects within the specified tolerance, reports and interprets a measure of statistical uncertainty (such as a confidence interval or standard error). Partial credit is given when only some of these elements are addressed. For instance, if the confounder is identified and a method applied but uncertainty quantification or a limitations discussion is omitted. No credit is awarded if mediator–outcome confounding is ignored, only unadjusted mediation estimates are presented, or methodological choices are left unjustified.

\paragraph{Sequential Mediators}

Sequential mediation occurs when an intervention exerts its effect on an outcome through a chain of two or more intermediate variables. The task requires disentangling each step in the mediator sequence: identifying how the treatment first influences one mediator, how that mediator then affects the next, and so on until the final outcome.

\textit{Motivation and Relevance:}  
Many real-world interventions operate via multiple stages. For example, an educational program that first enhances coping strategies, which in turn builds social support, ultimately improving mental health. Ignoring the sequential nature of these mediators can misattribute effects to the wrong pathway and lead to suboptimal or even counterproductive interventions. LLMs are especially challenged by this scenario because they must recognize the ordered dependencies among mediators and apply methods (such as path-specific effects analysis or structural equation models) rather than treating mediators as independent channels.

\textit{Data Generation:}  
We simulate five datasets reflecting a realistic educational intervention scenario:

\begin{itemize}[leftmargin=*, itemsep=1pt, topsep=3pt]
  \item Educational Intervention (Treatment): Whether students receive a skills-training program.
  \item Coping Strategy Improvement (Mediator 1): Increase in students’ coping skills following the intervention.
  \item Social Support Enhancement (Mediator 2): Growth in support networks resulting from improved coping.
  \item Mental Health Score (Outcome): Final assessment of students’ psychological well-being.
  \item Confounders: Socioeconomic status, school quality, and baseline cognitive score (each affecting treatment assignment and mediators).
\end{itemize}

The underlying causal structure is depicted by the DAG:

\begin{center}
\begin{tikzpicture}[every node/.style={
    draw,
    rectangle,
    rounded corners,
    align=center,
    inner sep=4pt
  }]
  \node (C) at (0, 0) {Confounders};

  \node (T) at (-5, -2) {Educational\\Intervention};
  \node (M1) at (-2, -2) {Coping Strategy};
  \node (M2) at ( 1, -2) {Social Support};
  \node (Y)  at ( 5, -2) {Mental Health Score};

  \draw[->] (C) -- (T);
  \draw[->] (C) -- (M1);
  \draw[->] (C) -- (M2);
  \draw[->] (C) -- (Y);

  \draw[->] (T) -- (M1);
  \draw[->] (T) to[bend right=25] (Y);
  \draw[->] (M1) -- (M2);
  \draw[->] (M2) -- (Y);
  \draw[->] (M1) to[bend right=15] (Y);
\end{tikzpicture}
\end{center}

\textit{Evaluation Rubric:}  
Models are evaluated on their ability to recover the sequential mediation structure and produce unbiased path-specific estimates. Full credit is awarded if the response correctly identifies all true confounders and both mediators in their proper order, excludes irrelevant variables, selects and justifies an appropriate sequential mediation method (for example, path-specific effects analysis or structural equation modeling), contrasts adjusted indirect effects with na{\"\i}ve estimates, provides numerical effect estimates within the specified tolerance, reports and interprets statistical uncertainty, and acknowledges observational limitations. Partial credit is given when only some of these elements are addressed (for instance, correctly identifying mediators but omitting uncertainty quantification). No credit is given if the model ignores the sequential pathway, applies only naïve mediation analysis, or fails to justify methodological choices.

\paragraph{Treatment–Mediator Interaction Effects}

Treatment–mediator interaction arises when the magnitude of a treatment’s effect on an outcome varies with the level of a mediator. In other words, the indirect pathway through the mediator modifies the direct effect of the treatment. Ignoring these interactions can mask important synergies or antagonisms, leading to incorrect estimates of direct and indirect effects.

\textit{Motivation and Relevance:}  
Accurately modeling treatment–mediator interactions is vital because policies or interventions may only be effective under specific mediator conditions. For example, an economic stimulus might only boost employment significantly when consumer confidence is already high. LLMs often default to additive causal models and overlook such interaction terms, risking misleading conclusions and suboptimal policy or clinical recommendations.

\textit{Data Generation:}  
We design five datasets to illustrate treatment–mediator interaction effects in an economic context. 

\begin{itemize}[leftmargin=*, itemsep=1pt, topsep=3pt]
  \item Stimulus Policy (Treatment): Indicates whether a region implements an economic stimulus (\textit{Yes}/\textit{No}).
  \item Consumer Confidence (Mediator): Public confidence level (\textit{High}/\textit{Low}), which may modify the policy’s effect.
  \item Employment Rate (Outcome): Regional employment outcome (\textit{Increased}/\textit{Not increased}).
  \item  Confounders: Market volatility and baseline economic growth, each affecting treatment assignment, consumer confidence, and employment.
  \item Irrelevant Variable: An unrelated indicator, included to test the model’s ability to exclude non-pertinent features.
  \item Interaction Term: The product of Stimulus Policy and Consumer Confidence,  included in the outcome model to capture effect modification.
\end{itemize}

The underlying causal structure is depicted by the DAG below:

\begin{center}
\begin{tikzpicture}[
  every node/.style={
    draw,
    rectangle,
    rounded corners,
    align=center,
    inner sep=4pt
  }
]
  \node (C) at (0, 0) {Market Volatility\\\& Baseline Growth};

  \node (T) at (-5, -2) {Stimulus Policy};
  \node (M) at ( 0, -2) {Consumer Confidence};
  \node (Y) at ( 5, -2) {Employment Rate};

  \draw[->] (C) -- (T);
  \draw[->] (C) -- (M);
  \draw[->] (C) -- (Y);

  \draw[->] (T) -- (M);
  \draw[->] (M) -- (Y);
  \draw[->] (T) to[bend right=15] (Y);
\end{tikzpicture}
\end{center}

\textit{Evaluation Rubric:}  
Models are evaluated on their ability to identify true confounders and the mediator, articulate how the treatment–mediator interaction modifies the causal effect, justify exclusion of irrelevant variables, select and justify an interaction-aware estimation method (such as regression with an interaction term), compare adjusted estimates to na{\"\i}ve associations, and provide numerical results with uncertainty measures. Full credit requires all these elements; partial credit is given when only some are addressed; no credit is awarded if interactions are ignored or methodological justifications are missing.

\subsection{Causal Discovery and Structure Learning}
Causal discovery and structure learning involve identifying unknown causal relationships and constructing comprehensive causal graphs from observational or experimental data. This type of analysis is distinct from tasks that estimate causal effects or perform mediation analysis. The primary goal here is to discover if causal links exist, determine their direction, and outline the overall causal structure from scratch. The process includes analyzing patterns of statistical dependence and conditional independence among variables. This can be particularly challenging for LLMs because they typically rely on correlations rather than deeper causal insights.

\paragraph{Cause–Effect Direction Inference}

Determining the correct causal direction between two correlated variables, deciding whether \(X_1\to X_2\) or \(X_2\to X_1\), is fundamental to causal discovery. This task requires more than observing associations; it demands identifying which variable truly generates the other, often by leveraging auxiliary information beyond the raw correlation.

\textit{Motivation and Relevance:}  
Inferring the wrong direction can lead to ineffective or harmful interventions, as policies and predictions depend critically on understanding which variable to manipulate. LLMs typically struggle here because they rely on surface-level correlations or heuristics. In practice, experts use additional clues (such as instrumental variables, temporal ordering, or distributional asymmetries) to distinguish cause from effect.

\textit{Data Generation:}  
We design five datasets in which two exogenous instruments (\(Z_1, Z_2\)) each influence exactly one of two target variables (\(X_1, X_2\)), and a single true causal edge runs between \(X_1\) and \(X_2\). An illustrative dataset is generated as follows:

\begin{itemize}[leftmargin=*, itemsep=1pt, topsep=3pt]
  \item \(Z_1, Z_2\) (\emph{Instruments}): Independent variables that affect \(X_1\) or \(X_2\) respectively, but not each other.
  \item \(X_1, X_2\) (\emph{Targets}): Two correlated variables linked by a true causal arrow (e.g.\ \(X_1\to X_2\)).
  \item Noise terms: Independent additive errors for each variable.
\end{itemize}

The causal structure for this illustrative dataset is depicted below:

\begin{center}
\begin{tikzpicture}[
  every node/.style={draw,rectangle,rounded corners,align=center,inner sep=4pt},
  node distance=1.8cm
]
  \node (Z1) at (-1,1) {\(Z_1\)};
  \node (Z2) at ( 1,1) {\(Z_2\)};
  \node (X1) at (-1,0) {\(X_1\)};
  \node (X2) at ( 1,0) {\(X_2\)};

  \draw[->] (Z1) -- (X1);
  \draw[->] (Z2) -- (X2);
  \draw[->] (X1) -- (X2);
\end{tikzpicture}
\end{center}

Here, each instrument \(Z_i\) is designed to have a nonzero effect only on the corresponding \(X_i\), and the single arrow \(X_1\to X_2\) represents the ground-truth causal direction.

\textit{Evaluation Rubric:}  
Models are evaluated on their ability to assign instruments correctly, carry out the two required regressions of \(X_1\) and \(X_2\) on \(\{Z_1,Z_2\}\), report $p$-values and identify which instrument is significant in only one regression, infer the causal arrow accordingly, and match that inference to the known truth. Full credit is awarded when all these elements are present; partial credit when some are addressed; and no credit if the solution relies only on correlations or omits any required component.

\paragraph{Handling Uncertainty in Causal Structures}

Handling uncertainty in causal structure inference requires acknowledging that, for a given set of observed conditional independencies, multiple Directed Acyclic Graphs (DAGs) may be equally compatible with the data. This phenomenon, known as Markov equivalence, means that one cannot definitively pinpoint a single causal graph without additional assumptions or interventions.

\textit{Motivation and Relevance:}  
Properly communicating uncertainty over plausible causal models is essential in fields such as epidemiology, economics, and social science, where decisions hinge on structural conclusions. Overconfident or overly definitive causal claims can mislead policy-makers or clinicians. While human experts often report equivalence classes or express confidence intervals over edges, LLMs tend to present a single ``best'' graph, failing to convey ambiguity or the need for further evidence.

\textit{Data Generation:}  
We sample five SEM datasets, and in each case, the true data‐generating DAG is a simple forward chain \(X_1\to X_2\to X_3\), and independent gaussian noise terms are added with no hidden confounding. The model is chosen that the forward chain and its two Markov-equivalent counterparts are indistinguishable by purely observational tests.

\begin{center}
\begin{tikzpicture}[node distance=20mm, every node/.style={
    draw,
    rectangle,
    rounded corners,
    align=center,
    inner sep=4pt
  }]
  \node (X1) {\(X_1\)};
  \node[right=of X1] (X2) {\(X_2\)};
  \node[right=of X2] (X3) {\(X_3\)};

  \draw[->] (X1) -- (X2);
  \draw[->] (X2) -- (X3);
\end{tikzpicture}
\end{center}

\textit{Evaluation Rubric:}  
Models are evaluated on whether they perform necessary marginal and conditional independence tests, list all Markov‐equivalent DAGs (including the forward chain, its reverse‐fork, and the reverse chain) without proposing any unsupported structures, and  refer to Markov equivalence or equivalence‐class ambiguity rather than presenting a single graph with unwarranted certainty. Full credit requires correct recognition and expression of uncertainty among all valid graphs; partial credit is given for acknowledging some but not all aspects of the equivalence class or uncertainty; no credit is awarded if the model reports a single graph as definitive or omits the equivalence consideration altogether.

\subsection{Causal Generalization and External Validity}
Causal generalization and external validity address the critical issue of whether a causal relationship identified in one context remains valid when applied to another context. Unlike internal validity, which confirms that observed effects are truly due to the intervention within a specific setting, external validity assesses the transferability of causal claims across varying conditions, populations, or environments. LLMs often assume learned causal relationships universally apply, failing to detect subtle contextual differences that could invalidate such assumptions. Robust causal reasoning thus demands recognizing the limits and scope of causal knowledge.

\paragraph{Population Shift and Transferability.}
 
Population shift arises when causal effects estimated in one group (e.g.\ Population A) do not generalize to another group (Population B) due to differences in baseline characteristics or effect modification. Transferability requires recognising when and how causal conclusions must be adjusted before being applied across populations.

\textit{Motivation and Relevance:}  
Failing to account for population shift can lead to ineffective or harmful interventions when policies validated in one demographic are indiscriminately exported to another. For example, an educational program that boosts test scores in urban schools may underperform or backfire in rural settings with different socioeconomic profiles. LLMs often extrapolate causal claims without acknowledging population nuances; robust methods are needed to ensure safe and effective transfer.

\textit{Data Generation:}  
We design five datasets comparing two populations (\textit{A} vs.\ \textit{B}). Each record includes:
\begin{itemize}[leftmargin=*, itemsep=1pt, topsep=3pt]
  \item Population: Indicator (\textit{A} or \textit{B}), determining which structural parameters apply.
  \item SocioEconomic Status: A baseline covariate influencing both program uptake and outcomes.
  \item Program Intensity: The ``dose'' of an educational intervention.
  \item Test Score: The measured outcome.
\end{itemize}

\begin{center}
\begin{tikzpicture}[node distance=18mm, every node/.style={
    draw,
    rectangle,
    rounded corners,
    align=center,
    inner sep=4pt
  }]
  \node (U) at (0,0) {Population \\ (\textit{A} vs \textit{B})};
  \node (SES) at (-5,-2) {SocioEconomic Status};
  \node (PI)  at (0,-2) {Program Intensity};
  \node (TS)  at (5,-2) {Test Score};

  \draw[->] (U)  -- (PI);
  \draw[->] (U)  -- (TS);
  \draw[->] (SES) -- (PI);
  \draw[->] (SES) to[bend right=15] (TS);
  \draw[->] (PI) -- (TS);
\end{tikzpicture}
\end{center}

\textit{Evaluation Rubric:}  
Models are evaluated on their ability to distinguish Population \textit{A} and \textit{B}, report the coefficient of Program Intensity on Test Score (with statistical significance or uncertainty) for each population, compare the magnitudes of these effects across populations, and summarize the implications for transferability. Full credit is awarded when all these elements are present and correctly interpreted; partial credit when only some are addressed; and no credit if the model ignores population distinctions or fails to compare and contextualize effect estimates across groups.

\paragraph{Temporal Stability of Causal Effects}

Temporal stability examines whether a causal effect remains constant across different time periods or whether it shifts due to changing conditions. The task requires detecting and quantifying any differences in the treatment’s impact on the outcome between an initial period and a later period.

\textit{Motivation and Relevance:}  
Evaluating temporal stability is vital for reliable forecasting and adaptive policy-making. An intervention that drives strong effects early on may weaken, or even reverse, its impact as underlying dynamics evolve. LLMs often assume static relationships, risking misleading guidance when effects drift. Experts counter this by stratifying analyses by time period, including interaction terms with time, or using time-varying coefficient models to capture and adjust for temporal change.

\textit{Data Generation:}  
We simulate five datasets illustrating temporal drift in a user-engagement scenario. Each record includes:

\begin{itemize}[leftmargin=*, itemsep=1pt, topsep=3pt]
  \item Time Period: Indicator of Period 1 or Period 2.
  \item User Invites (Treatment): Number of invitations sent to potential users.
  \item User Signups (Outcome): Number of users who sign up. 
\end{itemize}

The causal structure is depicted by the DAG below:

\begin{center}
\begin{tikzpicture}[node distance=20mm, every node/.style={
    draw,
    rectangle,
    rounded corners,
    align=center,
    inner sep=4pt
  }]
  \node (Time)    {Time Period};
  \node[right=of Time] (Invites) {User Invites};
  \node[right=of Invites] (Signups) {User Signups};

  \draw[->] (Time)    -- (Invites);
  \draw[->] (Time)    to[bend right=15] (Signups);
  \draw[->] (Invites) -- (Signups);
\end{tikzpicture}
\end{center}

\textit{Evaluation Rubric:}  
Models are evaluated on their ability to distinguish Period 1 from Period 2, fit separate analyses of signups on invites for each period, report the estimated effect sizes within the specified tolerance, discuss statistical significance, interpret whether the effect has strengthened or weakened over time, and conclude appropriately about temporal variation. Full credit is awarded when all these elements are present; partial credit if only some aspects are addressed; and no credit if temporal drift is ignored or treated as time-invariant.

\paragraph{Contextual Interaction and Moderation Effects}

Contextual interaction, or moderation, occurs when the effect of a treatment on an outcome depends on the level of another variable (the moderator). The task is to identify and quantify how this moderator alters the treatment’s impact, rather than assuming a uniform effect across all contexts.

\textit{Motivation and Relevance:}  
Accounting for moderation is critical because interventions may only be effective under particular conditions. For example, a pain medication might relieve symptoms in mild arthritis but be less effective or even counterproductive in severe cases. LLMs often overlook such nuances, treating effects as homogeneous and risking flawed recommendations. Experts  model interaction terms or conduct stratified analyses to capture these context-dependent effects.

\textit{Data Generation:}  
We design five datasets illustrating dosage–severity interactions in clinical and biological settings.

\begin{itemize}[leftmargin=*, itemsep=1pt, topsep=3pt]
  \item Condition Severity: A continuous or ordinal measure of baseline severity.
  \item Treatment Dosage: Dosage level administered.
  \item Symptom Reduction: The observed decrease in symptoms.
  \item Interaction Term: The product of Condition Severity and Treatment Dosage is  included in the outcome model to generate a moderation effect.
\end{itemize}

The causal structure is represented by the DAG below:

\begin{center}
\begin{tikzpicture}[node distance=20mm, every node/.style={
    draw,
    rectangle,
    rounded corners,
    align=center,
    inner sep=4pt
  }]
  \node (S) {Severity};
  \node[right=of S] (D) {Dosage};
  \node[below=of $(S)!0.5!(D)$] (Y) {Reduction};

  \draw[->] (S) -- (D);
  \draw[->] (S) -- (Y);
  \draw[->] (D) -- (Y);
\end{tikzpicture}
\end{center}

\textit{Evaluation Rubric:}  
Models are evaluated on their ability to report the main effect of dosage, the main effect of severity, and the interaction coefficient; to interpret how severity levels moderate the dosage effect; to perform and mention any diagnostic checks (e.g.\ residual analysis); and to summarize the implications for tailored dosing. Full credit is awarded when all elements are addressed with appropriate numerical or qualitative detail; partial credit when only some are; and no credit if moderation is ignored or methodological justifications are omitted.

\paragraph{Domain Shift and Transportability of Causal Knowledge}

Domain shift arises when causal conclusions drawn in one population fail to generalize to another due to differences in baseline characteristics or effect modifiers. Transportability concerns identifying these differences and adjusting causal inferences so that interventions remain valid across domains.

\textit{Motivation and Relevance:}  
Ignoring domain shift can lead to interventions that backfire or underperform when moved to new settings. For example, a drug shown to reduce blood pressure in middle-aged adults may be less effective in older patients with higher frailty. LLMs often extrapolate causally without accounting for such nuances, whereas experts employ statistical techniques to pinpoint and correct for population differences, ensuring safe and effective transfer of causal knowledge.

\textit{Data Generation:}  
We design five datasets contrasting two age-defined cohorts (Group 1: ages 30–50; Group 2: ages 50–70).

\begin{itemize}[leftmargin=*, itemsep=1pt, topsep=3pt]
  \item Age: Numeric age to determine the membership of the cohort.
  \item Frailty: A health indicator positively correlated with age.
  \item Drug Dose (Treatment): Dose of drug taken.
  \item Blood Pressure Reduction (Outcome): Measured drop in systolic blood pressure.
\end{itemize}
 This design forces models to grapple with shifting treatment effects under different demographic distributions.

\begin{center}
\begin{tikzpicture}[node distance=18mm,every node/.style={draw,rectangle,rounded corners,align=center,inner sep=4pt}]
  \node (A) at (-4,-2) {Age};
  \node (F) at (0,-2) {Frailty};
  \node (D) at (4,-2) {Drug Dose};
  \node (Y) at (0,-4) {BP Reduction};
  \draw[->] (A) -- (F);
  \draw[->] (F) -- (Y);
  \draw[->] (F) -- (D);
  \draw[->] (D) -- (Y);
  \draw[->] (A) -- (Y);
\end{tikzpicture}
\end{center}

\textit{Evaluation Rubric:}  
Models are evaluated on their ability to propose splitting the data into the correct age strata, specify and justify a linear model of blood pressure reduction on drug dose and frailty within each stratum, condition the estimate on the individual’s age scenario (e.g.\ age = 65), recognize frailty as a necessary input for precise effect estimation, explain how frailty confounds the dose–response in the older cohort, and request any additional information needed for a complete answer. Full credit is awarded when all elements are addressed; partial credit when only some are present; and no credit if domain differences are ignored or methodological steps are omitted.

\newpage
\section{Details of Two Protocols}

We evaluated models on causal inference tasks using two different prompting approaches: \textbf{Direct Prompting} and \textbf{Code Assisted Prompting}. Each method tests a distinct aspect of the model’s abilities.

\subsection{Protocol 1: Direct Prompting}

In the \textbf{Direct Prompting} protocol, we test the model's ability to analyze causal questions directly from raw data, without using computational code.

\begin{tcolorbox}[enhanced,
    colback=blue!5!white,
    colframe=blue!60!black,
    fonttitle=\bfseries\sffamily,
    title=Workflow for Direct Prompting,
    boxrule=0.4mm,
    arc=2mm,
    drop shadow=black!10!white,
    before skip=10pt,
    after skip=10pt]
\begin{enumerate}[leftmargin=*]
    \item \textbf{Model Prompting:}  
    Present the question alongside the raw data to the model.

    \item \textbf{Output Collection:}  
    Record the model's response for evaluation.
\end{enumerate}
\end{tcolorbox}

\textbf{Example:}

\begin{tcolorbox}[enhanced,
    colback=white,
    colframe=gray!70!black,
    boxrule=0.3mm,
    arc=2mm,
    drop shadow=black!10!white,
    before skip=10pt,
    after skip=10pt]
\small
\textit{"Question: Evaluate whether \{TREATMENT\} causally affects \{OUTCOME\}.\\[4pt]
Data: \{data\}\\[4pt]
Provide your analysis without using code."}
\end{tcolorbox}

This method focuses on the model’s ability to reason intuitively and draw conclusions directly from the provided data.

\subsection{Protocol 2: Code-Assisted Prompting}

The \textbf{Code-Assisted Prompting} protocol tests whether the model can identify causal issues, generate relevant Python code to analyze the data, and interpret the numerical outcomes to resolve the causal question.

\begin{tcolorbox}[enhanced,
    colback=green!5!white,
    colframe=green!50!black,
    fonttitle=\bfseries\sffamily,
    title=Workflow for Code Assisted Prompting,
    boxrule=0.4mm,
    arc=2mm,
    drop shadow=black!10!white,
    before skip=10pt,
    after skip=10pt]
\begin{enumerate}[leftmargin=*]
    \item \textbf{Code Generation:}  
    Provide the model with the causal question, dataset location, column names, and a small data sample (10 rows). Request Python code for analysis.

    \item \textbf{Code Execution:}  
    Extract and run the generated Python code to obtain numerical results.

    \item \textbf{Result Interpretation:}  
    Show the model the code it generated and its numerical results, and ask it to interpret these results in context.

    \item \textbf{Output Collection:}  
    Record the model's interpretation and analysis for evaluation.
\end{enumerate}
\end{tcolorbox}

\textbf{Example (Code Generation Prompt):}

\begin{tcolorbox}[enhanced,
    colback=white,
    colframe=gray!70!black,
    boxrule=0.3mm,
    arc=2mm,
    drop shadow=black!10!white,
    before skip=10pt,
    after skip=10pt]
\small
\textit{"Question: Evaluate whether \{TREATMENT\} causally affects \{OUTCOME\}.\\[4pt]
Dataset location: \texttt{/path/to/dataset.csv}\\[4pt]
Columns: \texttt{\{col1, col2, col3, ...\}}\\[4pt]
Data sample (first 10 rows): \emph{\{data\_sample\}}\\[4pt]
Provide Python code to perform this analysis."}
\end{tcolorbox}

\textbf{Example (Results Interpretation Prompt):}

\begin{tcolorbox}[enhanced,
    colback=white,
    colframe=gray!70!black,
    boxrule=0.3mm,
    arc=2mm,
    drop shadow=black!10!white,
    before skip=10pt,
    after skip=10pt]
\small
\textit{"Question: Evaluate whether \{TREATMENT\} causally affects \{OUTCOME\}.\\[4pt]
Code:\\[3pt]
\emph{\{Python code generated by the model\}}\\[4pt]
Results obtained:\\[3pt]
\emph{\{Numerical results from code execution\}}\\[4pt]
Based on these results, provide your analysis and answer the causal question."}
\end{tcolorbox}

Together, the two protocols provide complementary insights, evaluating both the intuitive reasoning and computational planning of models in causal inference tasks.

\section{Causal Pitfall Examples}
\label{app:two_causal_pitfalls}

This appendix provides detailed descriptions of two illustrative examples as introduced in Section~\ref{sec:two_pitfalls}.

\subsection{Branding Bias: Adversarial Sensitivity to Branding and Semantic Manipulation}

To highlight the susceptibility of LLMs to semantic manipulations, we designed an adversarial scenario testing whether beverage branding influences LLM-based causal conclusions about health impacts. The underlying causal structure was fixed, ensuring beverage consumption directly affected health outcomes positively or negatively, while lifestyle and health awareness independently affected both beverage consumption and health. Importantly, the beverage brand names (``HealthPlus'' or ``UltraSugar'') themselves had no actual causal effect (see Figure~\ref{fig:branding_bias_dag} in the main paper).

\paragraph{Dataset Construction.}
We generated two datasets to represent four distinct scenarios, each combining a brand name (``HealthPlus'', ``UltraSugar'') with a true health effect (beneficial or harmful):

\begin{enumerate}
\item  Brand ``HealthPlus'', truly beneficial effect.
\item  Brand ``HealthPlus'', truly harmful effect.
\item  Brand ``UltraSugar'', truly harmful effect.
\item  Brand ``UltraSugar'', truly beneficial effect.
\end{enumerate}

Each dataset included 200 samples with variables: Consumption, Outcome (health impact), Health Awareness, and Lifestyle. 

\paragraph{LLM Performance and Observations.}
LLMs were asked to assess if each beverage (``HealthPlus'' or ``UltraSugar'') was beneficial or harmful based purely on the given data. Table~\ref{tab:branding_bias_results} in the main paper summarizes the LLM conclusions across scenarios. 

\subsection{Spurious Causal Inference from Random Patterns.} 
We analyze the dataset \texttt{research\_funding\_rates} from the \texttt{dslabs} R package to illustrate how a rigorous causal analysis should be conducted.  

We first construct a dataset containing the number of applications, awards, and success rates for each gender. Disciplines are then re-ordered by their overall success rate from the original dataset. To examine whether the data exhibit Simpson’s paradox, we plot success rates by discipline (ordered by overall success), using color to indicate gender and point size to reflect the number of applications.  

\begin{figure}[ht]
    \centering
    \includegraphics[width=0.7\linewidth]{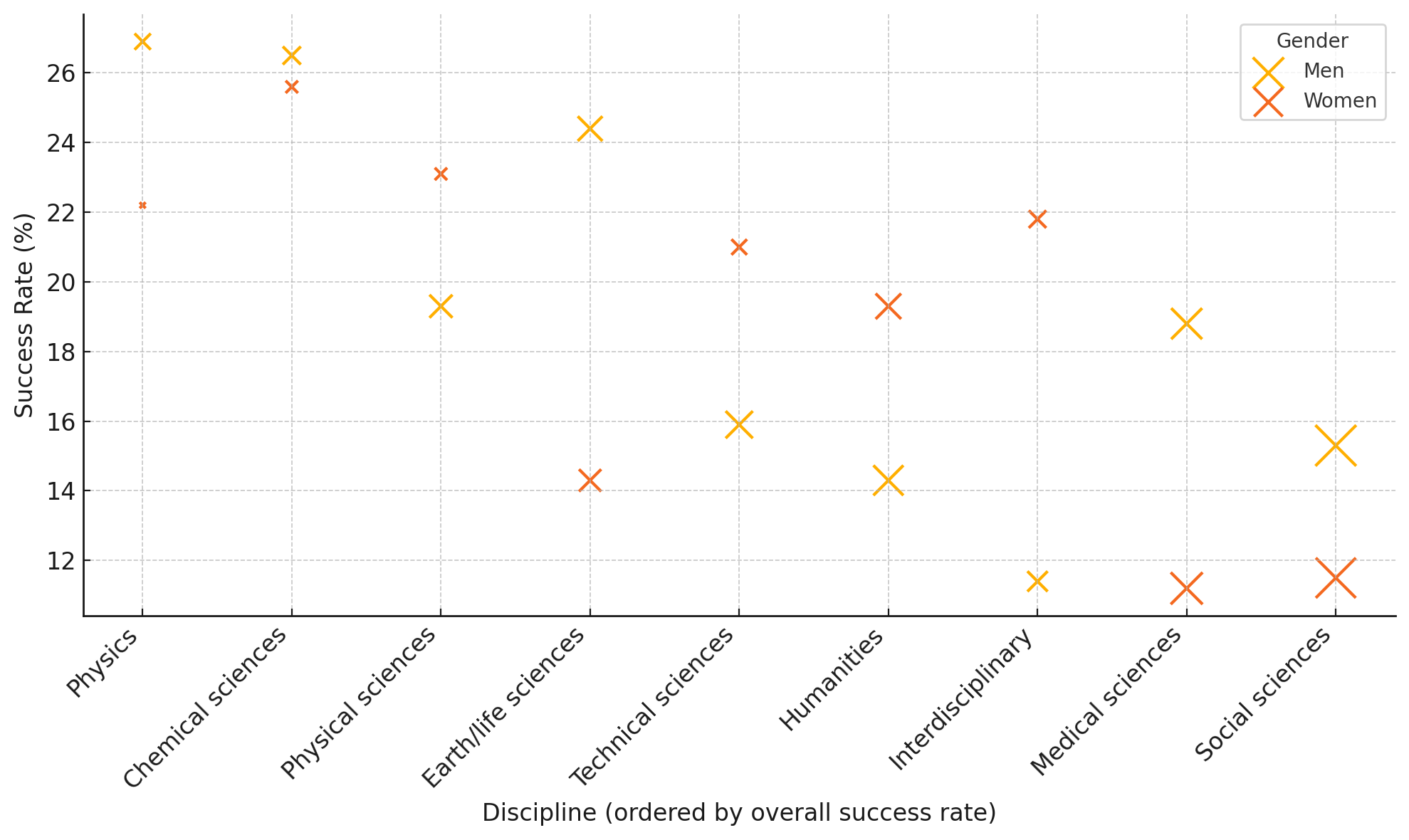}
    \caption{Success rates across disciplines for men and women.}
    \label{fig:Success_discipline}
\end{figure}

As shown in Figure~\ref{fig:Success_discipline}, there is no clear confounder driving the observed patterns. Nonetheless, some fields appear to favor men, while others favor women. Notably, the two disciplines with the largest differences favoring men are also those with the largest number of applications.  

This raises the question: could some selection committees be biased while others are not? To investigate, we compute the log-odds ratio divided by its standard error for each discipline and examine their distribution using a Q–Q plot against the standard normal distribution.  

\begin{figure}[ht]
    \centering
    \includegraphics[width=0.7\linewidth]{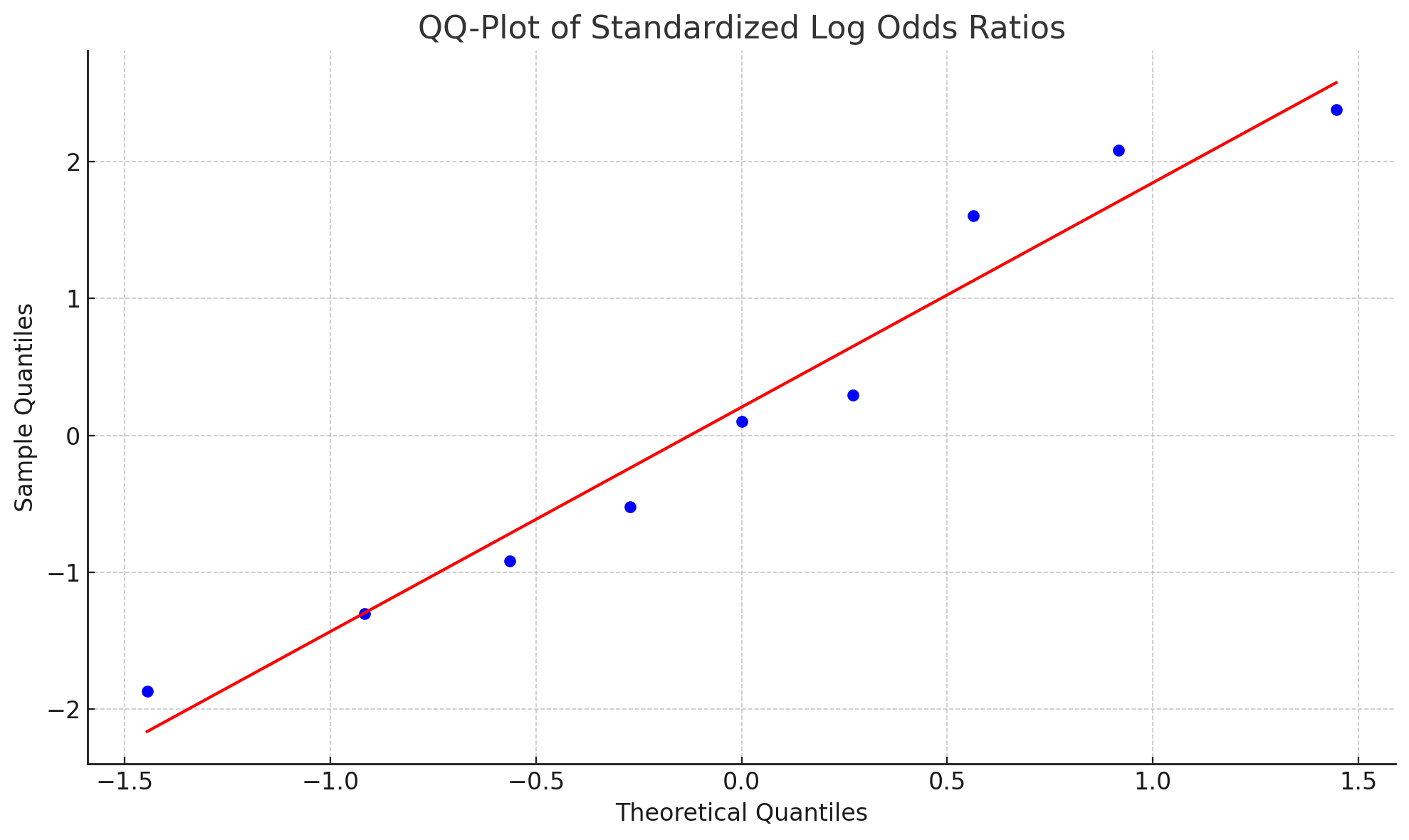}
    \caption{QQ-plot of standardized log odds ratios across disciplines.}
    \label{fig:qq_plot}
\end{figure}

Figure~\ref{fig:qq_plot} shows that the points lie close to the reference line, indicating that the observed effect sizes across disciplines are consistent with random variation under the null hypothesis of no systematic bias. The single “significant” result in Medical Sciences, which emerges under a chi-square test, is plausibly attributable to chance when conducting multiple comparisons, rather than reflecting genuine committee-level bias.

\section{Human-LLM Grading Alignment}

To assess the reliability of our GPT-4o-based automated evaluation approach, we conducted a validation involving expert human graders. This section elaborates on our detailed validation procedure.

\paragraph{Sampling Procedure.}
We randomly selected 150 LLM-generated responses from our complete evaluation dataset, employing \textbf{stratified} random sampling to ensure balanced representation across:
\begin{enumerate}
    \item First, we divided responses into the six causal inference categories, selecting an equal number (25) from each category.
    \item Within each category, we evenly sampled across the five difficulty levels (very easy, easy, medium, hard, very hard), selecting exactly 5 responses per difficulty level.
    \item For each category-difficulty combination, we randomly selected responses from the evaluated LLMs, ensuring proportional representation of all models' outputs.
\end{enumerate}

This sampling approach ensured the validation set accurately represented the complexity, diversity, and balanced coverage of our entire evaluation dataset.

\paragraph{Human Grading Procedure.}
Each sampled response was independently evaluated by three PhD students majoring in statistics. Prior to grading, the evaluators received detailed instructions on the grading rubrics to ensure consistency in scoring. Each grader assessed responses independently based on these rubrics, after which we aggregated their individual scores by taking the arithmetic mean. The average \textit{Gap} (absolute scoring differences) between human-expert and GPT-4o automated scores across causal inference categories, are summarized in Table~\ref{tab:gap_by_category}.

\begin{table}[ht]
\caption{Average \textit{Gap} (absolute scoring differences) between human graders and GPT-4o across causal inference categories (25 samples per category).}
\label{tab:gap_by_category}
\centering
\small
\renewcommand{\arraystretch}{1.2}
\setlength{\tabcolsep}{10pt}
\begin{tabular}{lc}
\toprule
\textbf{Causal Pitfall Category} & \textbf{Average Gap} \\
\midrule
Confounding biases and spurious associations & 0.08 \\
Interventions and experimental reasoning & 0.05 \\
Counterfactual reasoning and hypotheticals & 0.15 \\
Mediation and indirect causal effects & 0.17 \\
Causal discovery and structure learning & 0.15 \\
Causal generalization and external validity & 0.05 \\
\midrule
\textbf{Overall (all categories)} & \textbf{0.11} \\
\bottomrule
\end{tabular}
\end{table}

\section{Additional Experimental Setup}

Our experiments included the following ten large language models:  
Llama3.1-8b, Llama3.1-70b, Mistral-7b, Mixtral-8x22b, Claude-3.5-sonnet, Gemma2-9b, Gemini-2.0-flash, Deepseek-chat, GPT-4.1, and GPT-o4-mini. 

All experiments and analyses were conducted using a single NVIDIA A100 GPU. API interactions were handled using the official SDKs provided by each model vendor (OpenAI, Anthropic, Google, Deepseek). 

Hyperparameters were consistently set across models to ensure fair comparisons. The maximum token length was standardized at 1000 tokens for all models. GPT o4-mini was configured using the parameter \texttt{reasoning\_effort="medium"}.

\section{Additional Experimental Results}
Tables~\ref{tab:direct_prompting_individual} and \ref{tab:code_assisted_individual} provide comprehensive evaluations of LLM reliability at the level of individual causal inference challenges, under both direct and code-assisted prompting protocols.

\begin{table}[ht!]
\caption{
Direct Prompting: normalized scores (\%) of LLMs across all individual causal inference challenges.
}
\label{tab:direct_prompting_individual}
\centering
\setlength{\tabcolsep}{3pt}
\renewcommand{\arraystretch}{1.1}
\scalebox{0.8}{
\begin{tabular}{lccccccccccccccc}
\toprule
\textbf{LLM} & CE & CD & CI & CF & DS & MC & NS & OE & PS & SB & SM & SP & SU & TS & TM \\
\midrule
Gemma2-9b          & 38.50 &  2.40 &  2.00 & 15.43 & 11.33 & 25.00 &  2.21 & 22.94 &  0.00 &  4.00 & 19.50 & 24.00 & 16.57 &  2.86 &  5.00 \\
Llama3.1-8b        & 40.00 &  4.00 &  4.00 & 21.71 & 15.33 & 38.50 &  1.54 & 29.56 &  0.00 &  7.20 & 29.00 & 29.71 & 11.43 &  6.86 & 20.50 \\
Llama3.1-70b       & 37.00 &  7.20 &  7.33 & 20.00 & 16.67 & 43.00 &  3.87 & 34.83 &  5.60 &  7.20 & 36.50 & 28.00 & 15.43 & 15.43 & 21.50 \\
Mistral-7b         & 37.00 &  4.80 &  2.00 & 10.86 & 21.33 & 18.50 &  2.04 & 22.67 &  0.80 &  7.20 & 25.00 & 27.43 & 12.00 &  0.57 & 14.00 \\
Mixtral-8x22b      & 39.50 &  6.40 &  5.33 & 16.00 & 18.00 & 35.00 &  2.64 & 24.67 &  4.80 &  8.80 & 33.00 & 24.00 & 12.00 & 10.86 & 14.50 \\
Claude-3.5-sonnet  & 51.88 & 15.20 & 13.33 & 17.71 & 18.06 & 59.00 &  9.73 & 43.33 & 25.60 &  7.20 & 40.50 & 30.29 & 22.36 & 22.29 & 22.00 \\
Gemini-2.0-flash   & 43.00 &  9.60 & 10.67 & 20.00 & 25.69 & 60.50 &  9.87 & 30.76 & 11.20 &  6.40 & 48.50 & 33.71 & 20.41 & 12.57 & 31.00 \\
Deepseek-chat      & 58.00 & 20.80 & 19.33 & 24.00 & 25.33 & 80.50 &  5.80 & 46.83 & 40.00 & 15.20 & 48.00 & 36.57 & 20.88 & 30.29 & 33.00 \\
GPT-4.1            & 36.72 & 18.40 & 16.67 & 10.29 & 22.67 & 65.79 &  6.59 & 31.56 & 38.40 &  8.80 & 61.31 & 25.71 & 14.29 & 18.29 & 37.00 \\
GPT-o4-mini        & 50.62 & 40.80 & 32.00 & 25.14 & 28.67 & 83.00 & 16.51 & 39.79 & 54.40 & 36.00 & 52.50 & 46.86 & 33.14 & 62.86 & 37.50 \\
\midrule
\textbf{Average}   & 43.22 & 12.96 & 11.27 & 18.11 & 20.31 & 50.88 &  6.08 & 32.69 & 18.08 & 10.80 & 39.38 & 30.63 & 17.85 & 18.29 & 23.60 \\
\bottomrule
\end{tabular}
}
\begin{flushleft}
{\footnotesize 
CE: Causal effect estimation; CD: Cause-effect direction; CI: Contextual interaction; CF: Counterfactual prediction; DS: Domain shift.
MC: Mediator-outcome confounding; NS: Necessity and sufficiency; OE: Observational vs experimental; PS: Population shift; SB: Selection bias.
SM: Sequential mediators; SP: Simpson's paradox; SU: Structure uncertainty; TS: Temporal stability; TM: Treatment-mediator interaction.
}
\end{flushleft}
\end{table}

\begin{table}[ht!]
\caption{
Code-Assisted Prompting: normalized scores (\%) of LLMs across all individual causal inference challenges.
}
\label{tab:code_assisted_individual}
\centering
\setlength{\tabcolsep}{3pt}
\renewcommand{\arraystretch}{1.1}
\scalebox{0.8}{
\begin{tabular}{lccccccccccccccc}
\toprule
\textbf{LLM} & CE & CD & CI & CF & DS & MC & NS & OE & PS & SB & SM & SP & SU & TS & TM \\
\midrule
Gemma2-9b          & 21.50 &  6.40 & 32.67 &  9.71 &  6.00 & 18.12 &  1.73 & 15.50 &  1.60 &  7.20 & 32.00 &  8.57 &  6.86 & 33.71 & 18.00 \\
Llama3.1-8b        & 19.44 &  6.96 & 15.97 & 12.57 & 21.53 & 17.50 &  1.69 & 12.72 &  7.20 &  5.00 & 21.59 & 22.32 &  5.44 & 24.57 & 15.83 \\
Llama3.1-70b       & 23.81 & 16.80 & 33.33 & 16.07 & 23.33 & 37.50 &  3.80 & 22.00 & 40.00 & 12.80 & 19.50 & 34.29 & 11.69 &  4.00 & 18.50 \\
Mistral-7b         & 15.91 &  3.20 & 20.29 &  2.29 & 14.67 & 12.00 &  1.90 & 10.51 &  1.80 &  0.80 & 16.48 &  9.09 &  9.94 &  2.29 &  5.50 \\
Mixtral-8x22b      & 29.17 & 18.18 & 30.67 &  9.94 & 20.29 & 33.70 &  2.90 & 31.67 & 16.80 & 11.67 & 26.70 & 33.33 &  9.77 & 29.14 & 18.75 \\
Claude-3.5-sonnet  & 53.95 & 13.64 & 24.64 & 17.14 & 23.61 & 48.44 &  9.57 & 40.72 & 20.80 & 18.40 & 48.91 & 47.43 & 19.33 & 36.57 & 28.12 \\
Gemini-2.0-flash   & 50.00 & 14.40 & 54.67 & 21.14 & 25.69 & 53.00 & 10.52 & 34.17 & 24.80 & 18.40 & 40.50 & 56.00 & 19.39 & 46.29 & 33.00 \\
Deepseek-chat      & 48.96 & 28.00 & 46.00 & 20.00 & 26.00 & 61.00 &  5.37 & 48.44 & 56.80 & 18.40 & 47.50 & 58.86 & 20.78 & 53.71 & 32.29 \\
GPT-4.1            & 37.50 & 26.40 & 51.96 & 21.14 & 28.00 & 60.16 &  6.96 & 44.50 & 60.00 & 32.00 & 56.55 & 62.29 & 20.71 & 55.43 & 36.50 \\
GPT-o4-mini        & 50.62 & 22.61 & 50.72 & 22.98 & 18.06 & 68.48 & 15.75 & 53.24 & 69.17 & 48.00 & 44.50 & 76.00 & 31.58 & 64.88 & 38.50 \\
\midrule
\textbf{Average}   & 35.09 & 15.66 & 36.09 & 15.30 & 20.72 & 40.99 &  6.02 & 31.35 & 29.90 & 17.27 & 35.42 & 40.82 & 15.55 & 35.06 & 24.50 \\
\bottomrule
\end{tabular}
}
\begin{flushleft}
{\footnotesize 
CE: Causal effect estimation; CD: Cause-effect direction; CI: Contextual interaction; CF: Counterfactual prediction; DS: Domain shift.
MC: Mediator-outcome confounding; NS: Necessity and sufficiency; OE: Observational vs experimental; PS: Population shift; SB: Selection bias.
SM: Sequential mediators; SP: Simpson's paradox; SU: Structure uncertainty; TS: Temporal stability; TM: Treatment-mediator interaction.
}
\end{flushleft}
\end{table}

  \section{Code-Assisted Prompting with Debugging}
  \label{sec:debugging_protocol}

  To further investigate whether the performance gap under code-assisted prompting arises
  from unreliable code generation, we introduce a variant: \textbf{Code-Assisted
  Prompting with Debugging}. This protocol extends Protocol~2 by allowing the model one
  opportunity to correct its code when execution fails. Specifically, when the generated
  code raises an error, we present the error message to the model and request a fixed
   version and re-execute the code.

  \begin{tcolorbox}[enhanced,
      colback=orange!5!white,
      colframe=orange!60!black,
      fonttitle=\bfseries\sffamily,
      title=Workflow for Code-Assisted Prompting with Debugging,
      boxrule=0.4mm,
      arc=2mm,
      drop shadow=black!10!white,
      before skip=10pt,
      after skip=10pt]
  \begin{enumerate}[leftmargin=*]
      \item \textbf{Code Generation:}
      Same as Protocol~2: provide the causal question, dataset location, column names,
  and a data sample. Request Python code for analysis.

      \item \textbf{Code Execution:}
      Extract and run the generated code.

      \item \textbf{Debugging (if execution fails):}
      Present the error message to the model and request corrected code. Execute the
  corrected code.

      \item \textbf{Result Interpretation:}
      Show the model its code and the numerical results, and ask it to interpret the
  results in context.

      \item \textbf{Output Collection:}
      Record the model's interpretation for evaluation.
  \end{enumerate}
  \end{tcolorbox}

  Table~\ref{tab:debugging_comparison} compares causal reliability across all three
  protocols. Debugging primarily benefits models that frequently fail on the first code
  attempt. For example, Mistral-7B improves from $7.65\%$ to $17.55\%$ and Llama3.1-8B
  from $12.86\%$ to $19.04\%$, recovering to or above their direct-prompting baselines.
  In contrast, stronger models such as GPT-o4-mini and DeepSeek-chat show mild but
  consistent improvements, as their initial code-error rates are already low. This
  pattern suggests that for weaker models, code-assisted prompting introduces a
  code-generation failure mode that masks their underlying reasoning ability, whereas
  stronger models leverage computation to improve causal inference.

  \begin{table}[ht!]
  \caption{Causal reliability (\%) across three evaluation protocols. Causal reliability
  is the normalized score averaged across all six pitfall categories.}
  \label{tab:debugging_comparison}
  \centering
  \small
  \renewcommand{\arraystretch}{1.15}
  \begin{tabular}{lccc}
  \toprule
  \textbf{LLM} & \textbf{Direct} & \textbf{Code-Assisted} & \textbf{Code-Assisted +
  Debug} \\
  \midrule
  Gemma2-9B          & 13.69 & 13.24 & 15.93 \\
  Llama3.1-8B        & 17.86 & 12.86 & 19.04 \\
  Llama3.1-70B       & 19.78 & 19.89 & 20.42 \\
  Mistral-7B         & 14.43 &  7.65 & 17.55 \\
  Mixtral-8$\times$22B & 16.99 & 20.59 & 20.84 \\
  Claude-3.5-Sonnet  & 26.22 & 29.38 & 30.17 \\
  Gemini-2.0-Flash   & 24.36 & 31.80 & 33.30 \\
  DeepSeek-Chat      & 32.43 & 36.12 & 36.72 \\
  GPT-4.1            & 25.24 & 37.32 & 37.98 \\
  GPT-o4-Mini        & 40.72 & 43.03 & 45.15 \\
  \bottomrule
  \end{tabular}
  \end{table}

\section{Benchmark Coverage Comparison}                                                
  \label{sec:benchmark_coverage}                                                         
                                                                                         
  Table~\ref{tab:benchmark_coverage} compares the challenge coverage of CausalPitfalls   
  against four existing benchmarks. Existing evaluations have focused either on semantic 
  causal reasoning~\citep{jin2023cladder} or on estimation accuracy in standard
  settings~\citep{liu2024llms,wang2024causalbench}. In contrast, CausalPitfalls targets
  common statistical pitfalls and failure modes across all six categories.

  \begin{table}[ht!]
  \caption{Coverage of CausalPitfalls challenges across existing benchmarks. ``Symbolic
  only'' refers to semantic reasoning without tabular data.}
  \label{tab:benchmark_coverage}
  \centering
  \small
  \renewcommand{\arraystretch}{1.15}
  \scalebox{0.85}{
  \begin{tabular}{lccccc}
  \toprule
  \textbf{Challenge} & \textbf{CLadder} & \textbf{Corr2Cause} & \textbf{CausalBench} &
  \textbf{QRData} & \textbf{Ours} \\
  \midrule
  Simpson's paradox & Symbolic only & No & No & Partial & Yes \\
  Selection bias (Berkson's) & Symbolic only & No & No & No & Yes \\
  Observational vs experimental & Symbolic only & No & Yes & Partial & Yes \\
  Causal effect estimation & Symbolic only & No & Yes & Partial & Yes \\
  Counterfactual prediction & Symbolic only & No & Yes & Limited & Yes \\
  Causal necessity \& sufficiency & Symbolic only & No & Partial & No & Yes \\
  Mediator--outcome confounding & Symbolic only & No & Partial & No & Yes \\
  Sequential mediators & Symbolic only & No & No & No & Yes \\
  Treatment--mediator interaction & No & No & No & No & Yes \\
  Cause--effect direction & Symbolic only & Symbolic only & Yes & No & Yes \\
  Uncertainty in causal structures & Partial & No & Partial & No & Yes \\
  Population shift \& transferability & No & No & No & No & Yes \\
  Temporal stability & No & No & No & No & Yes \\
  Contextual interaction \& moderation & No & No & Partial & Partial & Yes \\
  Domain shift \& transportability & No & No & No & No & Yes \\
  \bottomrule
  \end{tabular}
  }
  \end{table}

\end{document}